\documentclass[3p]{elsarticle}

\usepackage{lineno,hyperref}
\usepackage{graphicx}
\usepackage{subcaption}
\usepackage{mwe}

\journal{Information Sciences}

\usepackage{amsmath}

\usepackage[normalem]{ulem}
\useunder{\uline}{\ul}{}

\usepackage{algorithm} 
\usepackage{algorithmic} 

\usepackage{xcolor}

\newcommand{\todo}[1]{\textcolor{black}{#1}}
\newcommand{\revtwo}[1]{\textcolor{black}{#1}}
\usepackage{adjustbox}
\usepackage{bibentry}

\usepackage{amsfonts}
\usepackage{lscape}

\begin{document}

\begin{frontmatter}

\title{Balancing Pareto Front exploration of Non-dominated Tournament Genetic Algorithm (B-NTGA) in solving multi-objective NP-hard problems with constraints}


\author{Micha{\l} Antkiewicz}
\author{Pawe{\l} B. Myszkowski\corref{cor2}}

\address{Wroc{\l}aw University of Science and Technology}
\address{Faculty of Information and Communication Technology}
\address{ul.Ignacego {\L}ukasiewicza 5, 50-371 Wroc{\l}aw, Poland}

\cortext[cor2]{Corresponding author: pawel.myszkowski@pwr.edu.pl}

\begin{abstract}
The paper presents a new balanced selection operator applied to \revtwo{the proposed} Balanced Non-dominated Tournament Genetic Algorithm (B-NTGA) that actively uses archive to solve multi- and many-objective NP-hard combinatorial optimization problems with constraints. The primary motivation is to make B-NTGA more efficient in exploring Pareto Front Approximation (PFa), focusing on ''gaps'' and reducing some PFa regions' sampling too frequently. Such a balancing mechanism allows B-NTGA to be more adaptive and focus on less explored PFa regions. The proposed B-NTGA is investigated on two benchmark multi- and many-objective optimization real-world problems, like Thief Traveling Problem and  Multi-Skill Resource-Constrained Project Scheduling Problem. The results of experiments show that B-NTGA has a higher efficiency and better performance than state-of-the-art methods.   
\end{abstract}


\end{frontmatter}


\section{Introduction}
\label{sec:introduction}

In the paper, two real-world combinatorial practical NP-hard problems are investigated to verify the effectiveness of the proposed Balanced Non-dominated Tournament Genetic Algorithm (B-NTGA) -- Multi-Skill Resource-Constrained Project Scheduling Problem (MS-RCPSP) \cite{benchmar2015, benchmark2022} and Travelling Thief Problem (TTP) \cite{TTP}. Both are NP-hard, constrained, combinatorial, and require multi-objective optimization. Moreover, each of the above problems combine multiple subproblems (i.e., allocation, scheduling), and both subproblems have objectives that are interwoven, and finding an optimal solution for them does not provide the optimal solution to the whole problem. This specifically makes problems closer to applications of real-world requirements.

In multi- and many-objective optimization MS-RCPSP final schedule is evaluated by cost and duration. The MS-RCPSP is a real-world many-objective practical problem and belongs to the class of strongly NP-hard problems \cite{HARTMANN20221} with combinatorial solution space. The primary goal is to assign given tasks to limited resources and then place them on a timeline. Additional challenge comes with a set of constraints connected to skill requirements, their availability, and tasks' precedence relation. The schedule is evaluated across up to 5 objectives \cite{benchmark2022}: cost, duration, average cash flow, skill overuse, and average use of resources. It also consists of task-to-resource assignments and time-slot assignments. It does not represent all the aspects of real-world project scheduling but provides a good approximation of a number of challenges. 
The Travelling Thief Problem (TTP) \cite{TTP} problem was initially defined in 2013, and its practical and interesting character makes it highly popular -- GoogleScholar returns hundreds of papers as a result. The TTP problem connects two well-known problems: the Traveling Salesman Problem (TSP) and the Knapsack Problem (KNP). The multi-objective TTP solution must satisfy constraints (e.g., the capacity of knapsack) and combines two well-known NP-hard problems with interconnected objectives -- TSP and KNP. 
It is worth mentioning that both (TTP and MS-RCPSP) multi-objective optimization problems are NP-hard, constrained, and real-world. That means a large solution landscape cannot be searched algorithmically (only a meta-heuristic could be effective), and some solutions are infeasible (as it breaks some constraints and cannot be used in practice). We even do not know the ''best'' solution; because of the size of the solution landscape and the multi-objective character of the problem where the \emph{domination} relation exists between solutions. In this situation, we can only explore the solution landscape to build a \emph{Pareto Front approximation} (PFa) (includes all found non-dominated solutions) and compare the results of the studied methods. That makes solving problem(s) particularly demanding but also challenging and valuable for real-world applications.

TTP and MS-RCPSP problems are recent real-world problems, and some solving methods are presented in the literature. Some of them could be listed as follows.
In work \cite{hantco}, a hybridized Ant Colony Optimisation has been applied to solve MS-RCPSP as a single objective optimization problem (as a weighted sum of objectives). Also, Differential Evolution in \cite{degr} is applied to solve the MS-RCPSP problem effectively. Other metaheuristics are used like Knowledge-Based Fruit Fly Optimisation\cite{fruit} or Teaching-Learning Optimisation \cite{tlbo} or Genetic Programming Hyper-heuristic \cite{hyper}. Some methods solve the MS-RCPSP problem as a bi-objective problem, for example, Non-Dominated Tournament Genetic Algorithm 2 (NTGA2) \cite{NTGA2}, multi-objective Fruit Fly Optimization Algorithm \cite{fruit-mmo} or decomposition-based multi-objective genetic program. Also, hyper-heuristics \cite{hyper2021} is used to solve MS-RCPSP as a multi-objective problem. To our best knowledge, only two publications (e.g.\cite{NTGA2}\cite{benchmark2022}) present methods that solve many-objective MS-RCPSP using 5-objectivities. In \cite{NTGA2} as reference method applied to MS-RCPSP are: U-NSGA-III, $\theta$-DEA and NTGA2. Results show that NTGA2 is the most effective and scalable method. 
The latest survey of referenced methods applied to MS-RCPSP is presented in \cite{benchmark2022}, and as problem MS-RCPSP is wider scheduling context presented in surveys, e.g. \cite{snasel-surv}\cite{HARTMANN20221}. 
  
This paper presents a Balanced Non-dominated Tournament Genetic Algorithm (B-NTGA) method that bases of NTGA2\cite{NTGA2} in the context of a successful application to MS-RCPSP and TTP. The main NTGA2 advantage is a GAP selection that provides NTGA2 with the opportunity to examine the \emph{``low density``} regions of PFa during the evolution -- motivation for this strategy is that we get more ‘‘dense" PFa and potentially better not dominated points. Such a strategy uses phenotype distance between individuals to improve the evolution process. Some of the introduced mechanisms maintained the original concept, just to mention a few - Niching\cite{ref:niching}, Speciation\cite{ref:speciation} or Crowding Distance\cite{ref:crowding} focuses on distances in solution space. Moreover, hybrid selection operators \cite{hybrid-selection} in the literature link decomposition, domination, and indicator-based approach to effectively solve multi- and many-objective optimization problems. The operator GAP \cite{NTGA2} works similarly and focuses on ‘‘gaps" in PFa.

\revtwo{The main novelty of the proposed B-NTGA is inspired by the \emph{Upper Confidence Bound} (UCB)\cite{upper} formula which is not a standard element of evolutionary computation. It makes the backbone of the Monte Carlo Tree Search (MCTS)\cite{browne2012survey} algorithm to support a balance between exploring the search space and exploiting promising solutions, commonly used in game-playing and robotics. Authors of The Monte Carlo Elites \cite{sfikas2021monte} applied the UCB formula to balance the search in the behavior space -- solutions mapped to another space based on their behavior - e.g. angular velocity in robotics design.
In this paper, we incorporated the idea into the objective space to balance the PF selection, yet the biggest novelty is combining it with the density metric in a single formula. The density metric (such as Crowding Distance mentioned before) is commonly used to promote the exploration of search space with lower solution density. However, the standard static approach promotes the same regions until density changes - which might never happen due to natural constraints of the search space. Our proposed approach - the balanced GAP selection operator (which is density-based) promotes the least explored regions of PF while taking into account the frequency of their selection.}

The proposed B-NTGA actively uses \emph{archive} and extends these mechanisms due to the frequency of PFa probing and allows to focus on less explored regions but depending on the already used ‘‘budget" in a given region. B-NTGA is investigated to examine how effective balanced GAP selection determines potential GAP improvement. We propose B-NTGA to make exploration/exploitation more balanced (as some Pareto fronts regions are selected too often), adapted (boundary points monopolize selection), and dynamic (selection uses ‘‘budget" in each generation to reduce exploration already probed regions). 

The \textbf{main contribution} of this paper is proposed B-NTGA that actively uses \emph{archive} to explore/exploit current PFa -- some regions are dense, some have \emph{low density} including \emph{''gaps''}. The B-NTGA searches such regions, focusing on \emph{''gaps''}, the biggest potential to find new non-dominated solutions. The novelty of B-NTGA is that the method actively uses \emph{archive} and remembers regions visited too often in the evolution process -- then it reduces exploration in such regions in following generations. Thus, B-NTGA \emph{balances} on searching potentially interesting PFa regions (many ''gaps'') but also reduces exploration of such regions if there are no results, i.e., new non-dominated solution. Such a mechanism is very useful, particularly in searching solution space with many constraints -- as a potential interesting ''gap'' in PFa could include only infeasible solutions \todo{(i.e. some constraints are not met)}, and exploration of such solution search space is meaningless and time-consuming. B-NTGA reducing selection pressure in that situation selects other regions in the evolution process. Therefore, such a mechanism allows B-NTGA to be more efficient. Also, B-NTGA basing NTGA2 could reduce some elements (e.g., clone reduction, tournament selection as second selection); Thus, B-NTGA is less time-consuming, uses fewer parameters, and is more efficient than NTGA2 \todo{and investigated state-of-the-art methods}. Results of experiments showed that the proposed B-NTGA, in comparison to state-of-the-art methods, is highly effective for multi-objective optimization problems (like TTP and MS-RCPSP) and many-objective MS-RCPSP, where 5-objectives are used in optimization.

The rest of the article is structured as follows. In section 2 a short related work is given. The investigated MS-RCPSP and TTP problems are shortly defined in section 3. Motivation and proposed approach are given in section 4. Section 5 includes experimental results of proposed B-NTGA multi- and many-objective optimization. Lastly, the paper is concluded in section 6.

\section{Related Work}
\label{sec:related-work}

\todo{To demonstrate the overall performance of the proposed B-NTGA in solving NP-hard multi-objective optimization problems, we investigate B-NTGA by solving two selected NP-hard problems}: TTP and MS-RCPSP. Those problems exist in the literature, where various propositions to solve those problems are given. Mainly, some methods ignore multi-objective problem (MOP) nature and try to solve them as a single objective optimization problem. Others consider multi-objective defining dominance relation in solution landscape, others use indicators to deal with multi-objective optimization, but some use decomposition to solve the problem. In this section, some of the state-of-the-art and best-known discrete optimization methods to solve NP-hard problems are given: classic NSGA-II (see sec.\ref{sec:nsgaii}), U-NSGA-III (sec.\ref{sec:unsgaiii}), $\theta$-DEA (sec. \ref{sec:thetadea}), NTGA2 (sec.\ref{sub:ntga2}), SPEA2 (sec.\ref{sec:spea2}) and MOEA/D (sec.\ref{sec:moead}).
\todo{In the subsequent part of this section, alternative approaches to traditional evolutionary methods for balancing exploration and exploitation in solution space are described (sec.\ref{sec:exp}). Towards the end, the impact of the many-objectiveness on the solution space is presented (sec.\ref{sec:manyobj}), followed by a summary.}

Moreover, all researched metaheuristics to get values of objectivities of a given $feasible$ solution (i.e., schedule) in MS-RCPSP an extra procedure (\emph{Schedule Builder} based on greedy algorithm \cite{NTGA2} -- details presented in sec.\ref{sec:greedy}) must be used. However, in TTP, the values of the objectives are calculated by the final path and knapsack content.

\subsection{Non-Dominated Sorting Genetic Algorithm II}\label{sec:nsgaii}

The classical NSGA-II \cite{nsgaii} \cite{blank2017solving} \cite{snasel-surv} proposed in the year 2002 for multi-objective optimization in literature but is still present \cite{snasel-surv} and effectively applicable many problems, also for MS-RCPSP \cite{NTGA} or TTP \cite{blank2017solving}. NSGA-II (see Pseudocode \ref{alg:nsgaii}) is based on the genetic algorithm schema.

\begin{algorithm}[!ht]
\caption{Pseudocode of NSGA-II \label{alg:nsgaii} \cite{blank2017solving}}
\begin{algorithmic}[2]
\STATE $P_{current} \gets generateInitialPopulation()$
\STATE $evaluate(P_{current})$
\STATE $nonDominatedSorting(P_{current})$
\FOR { $i \gets 0$ to $generationLimit$ }
    \STATE $Q \gets createOffspringPopulation(P_{current})$
    \STATE $P_{current} \gets P_{current} \cup Q$
    \STATE $nonDominatedSorting(P_{current})$
    \STATE $truncate(P_{current})$
\ENDFOR
\end{algorithmic}
\end{algorithm}

 Initially, population initialization is performed (Pseudocode \ref{alg:nsgaii}, line 1), followed by the evaluation of individuals (line 2). For the MS-RCPSP, it is important to note that in order to obtain a \emph{feasible} schedule, a dedicated \emph{Schedule Builder} \cite{NTGA2} must be utilized. This \emph{Schedule Builder} aids in generating valid schedules that adhere to the constraints of the MS-RCPSP. For the TTP the values of the objectives are calculated by the final path and knapsack content. Next, the population in NSGA-II is sorted (line 3) based on their $rank$ and $crowding$ $distance$. The $rank$ signifies the number of the (Pareto) front to which an individual belongs. The $crowding$ $distance$ measures the volume of the smallest box that exclusively encloses a given individual within the front.

 The evolution begins in line 4 (see Pseudocode \ref{alg:nsgaii}), where an offspring population $Q$ is generated using genetic operators. For selection, the tournament selection method is employed, comparing the ranks of individuals. In cases where their ranks are equal, the crowding distance becomes the decisive factor. After the creation of the offspring population, both the parent and offspring populations (as seen in line 6) are sorted and then truncated (line 8) to their original size, ensuring that the population remains within the desired limit.

\subsection{Unified NSGA-III}
\label{sec:unsgaiii}

The U-NSGA-III \cite{u-nsgaiii} method is an extension of classic NSGA-II but differs in the main evolution loop (line 5). Moreover,  U-NSGA-III uses a set of widely distributed reference points \cite{referepoints}. In Pseudocode \ref{alg:unsgaiii} line 9 the non-dominated sorting of the population is made, then (Pareto) fronts $F_i$ are added to the next population. If the population exceeds the given size, a niching method is used to select individuals. First, all objectives are normalized in the next population and that front. 

\begin{algorithm}[!ht]
\caption{Pseudocode of U-NSGA-III \cite{u-nsgaiii} \label{alg:unsgaiii}}
\begin{algorithmic}[2]
\STATE $N \gets $ population size
\STATE $H \gets $ reference points
\STATE $P_{current} \gets generateInitialPopulation()$
\STATE $evaluate(P_{current})$
\FOR { $i \gets 0$ to $generationLimit$ }
    \STATE $P_{next} \gets $ empty array
    \STATE $Q \gets createOffspringPopulation(P_{current})$
    \STATE $P_{current} \gets P_{current} \cup Q$
    \STATE $(F_1, F_2, ...) \gets nonDominatedSorting(P_{current})$
    \STATE $i = 1$
    \WHILE { $|P_{next}| + |F_{i}| < N$ }
        \STATE $P_{next} \gets P_{next} \cup F_i$
        \STATE $i \gets i + 1$
    \ENDWHILE
    \STATE $normalize(F_{i}, P_{next}, H)$
    \STATE $associate(F_{i}, H)$
    \STATE $P_{next} \gets P_{next} \cup niching(F_{i})$
\ENDFOR
\end{algorithmic}
\end{algorithm}

Then, each individual of the front is associated (see line 16 in Pseudocode \ref{alg:unsgaiii}) with the closest reference point -- the perpendicular distance to the line defined by the reference points and the origin is used. Lastly, in line 17 individuals with the least associated individuals are selected for the next generation. If there are multiple individuals, the one closest to the reference line is selected.

\subsection{$\theta$-DEA}
\label{sec:thetadea}
The $\theta$-DEA \cite{thetadea} uses $\theta$-dominance relation, which associates each individual to two values: $d_1$ and $d_2$. Where $d_1$ is the distance to the $Perfect$ $Point$ (defined in sec.\ref{sec:experiments} in Quality Measure section) along one of the reference lines, and $d_2$ is the distance to the closest reference line.

\begin{algorithm}[!ht]
\caption{Pseudocode of $\theta$-DEA \cite{thetadea} \label{alg:thetadea}}
\begin{algorithmic}[2]
\STATE $N \gets $ population size
\STATE $H \gets $ reference points
\STATE $P_{current} \gets generateInitialPopulation()$
\STATE $evaluate(P_{current})$
\FOR { $i \gets 0$ to $generationLimit$ }
    \STATE $P_{next} \gets $ empty array
    \STATE $Q \gets createOffspringPopulation(P_{current})$
    \STATE $P_{current} \gets P_{current} \cup Q$
    \STATE $(F_1, F_2, ...) \gets nonDominatedSorting(P_{current})$
    \STATE $i = 1$
    \WHILE { $|P_{next}| + |F_{i}| < N$ }
        \STATE $P_{next} \gets P_{next} \cup F_i$
        \STATE $i \gets i + 1$
    \ENDWHILE
    \STATE $normalize(F_{i}, P_{next}, H)$
    \STATE $cluster(F_{i}, H)$
    \STATE $(F_1, F_2, ...) \gets thetaNonDominatedSorting(P_{current})$
    \STATE $i = 1$
    \WHILE { $|P_{next}| + |F_{i}| < N$ }
        \STATE $P_{next} \gets P_{next} \cup F_i$
        \STATE $i \gets i + 1$
    \ENDWHILE
    \STATE $P_{next} \gets P_{next} \cup addRandom(F_i)$
\ENDFOR
\end{algorithmic}
\end{algorithm}

The $\theta$-DEA is presented in Pseudocode \ref{alg:thetadea}, and its schema is similar to U-NSGA-III. But there is the main difference -- the last front is added to the population. In line 16, the clustering adds each individual of the front to the closest reference line and is sorted by $\theta$-dominance. The resulting $F_1$ contains all the solutions that are closest to their corresponding reference lines, where the distance $D$ is calculated as follows:

\begin{equation}\label{eq:theta-dist}
D = d_1 + \theta d_2
\end{equation}

Next, $F_2$ contains the (second) closest solutions, and so on. The individuals from the $F_i$ to the next population are added in a loop in lines 19-22. If merged front $F_i$ exceeds the population size, the random individuals (line 23) from given front $F_i$ are selected.

\subsection{Non-Dominated Tournament Genetic Algorithm 2}
\label{sub:ntga2}

NTGA2\cite{NTGA2} serves as a direct predecessor to the method introduced in this paper. It operates as an evolutionary technique that emphasizes diversity by utilizing a Gap Selection ($GS$) operator. $GS$ functions within the objective space, prioritizing regions of the archive that have been less explored. A detailed description of the $GS$ operator can be found below. NTGA2 utilizes an $archive$ to store all already found non-dominated solutions. This archival approach allows the method to maintain a diverse set of solutions that represent the non-dominated front and contribute to the optimization process.

\begin{algorithm}[!h]
\caption{Pseudocode of NTGA2 \cite{NTGA2}
\label{alg:ntga2}}
\begin{algorithmic}[1]
\STATE $archive \gets \emptyset$
\STATE $P_{current} \gets GenerateInitialPopulation()$
\STATE $Evaluate(P_{current})$
\STATE $UpdateArchive(P_{current})$
\FOR { $i \gets 0$ to $Generations$ }
    \STATE $P_{next} \gets \emptyset$
    \WHILE { $|P_{next}| < |P_{current}|$ }
        \IF { $i$ mod (2 * $gsGenerations) < gsGenerations$ }
            \STATE $Parents \gets Select_{tour}(P_{current})$
        \ELSE
            \STATE $Parents \gets Select_{GS}(P_{current} \cup archive)$
        \ENDIF
        \STATE $Children \gets Crossover(Parents)$
        \STATE $Children \gets Mutate(Children)$
        \WHILE { $P_{next}$ contains $Children$ }
            \STATE $Children \gets Mutate(Children)$
        \ENDWHILE
        \STATE $Evaluate(Children)$
        \STATE $P_{next} \gets P_{next} \cup Children$
        \STATE $UpdateArchive(Children)$
    \ENDWHILE
    \STATE $P_{current} \gets P_{next}$
\ENDFOR
\end{algorithmic}
\end{algorithm}

\begin{figure}[!ht]
\centering
\includegraphics[width=0.74\textwidth]{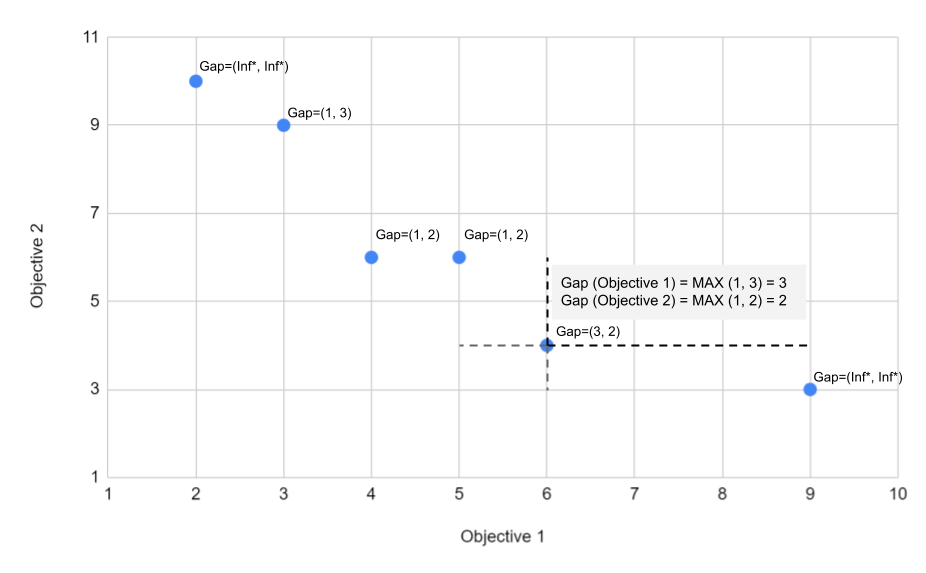}
\caption{The example of the ‘‘gap'' calculation for two objectives, used by the $Gap$ $Selection$ ($GS$) }
\label{fig:gap-scheme}
\end{figure}

The NTGA2 uses a random population $P_{current}$ initialization -- see the pseudocode \ref{alg:ntga2}, line 2. Then, all individuals are evaluated (separately by each objective) and then $UpdateArchive$ takes place, where non-dominated individuals are added and just dominated ones are removed.
The general NTGA2 loop starts at line 5 and lasts for a set number of $Generations$. Each generation starts with a selection of individuals to the new population $P_{next}$. $GS$, and the second selection (Pareto-dominance tournament selection) is used. The $gsGenerations$ parameter (line 8) selection decides which operator is used -- it is realized as a number of generations that has to pass before selection switches. Line 15 presents the clone elimination mechanism applied in NTGA2. The genetic operators (e.g. mutation and crossover) should be specialized per problem but can be set by default to standard single-point crossover and random bit mutation if it fits the problem encoding.
\\
\\
\textbf{Gap Selection} ($GS$) operator\cite{NTGA2} aims to increase the \emph{diversity} of PFa. It operates in an objective space and considers each objective separately. The authors decided to select objectives as follows: offspring generation is divided into $M$ parts (where $M$ is the number of objectives) and each objective is selected during the corresponding part. It starts by calculating the ‘‘gap" size for each individual in the $PFa$ (PFa). It is calculated considering the two neighbor individuals (minimal \emph{Euclidean} distance). Those are the closest individuals, one with a worse objective value and one with a better value. The $GapValue$ is used as the \emph{Euclidean} distance to the farther of those two neighbors. Additionally, individuals at the edge (highest and lowest values) of the approximation have this distance set to an infinity value. Fig. \ref{fig:gap-scheme} presents the calculation with an example. This mechanism can be compared to the $crowding$ $distance$\ref{alg:nsgaii} but differs in two aspects. The maximum neighbor distance is used, where $crowding$ $distance$ is based on the average. What is more important, $GS$ works in one dimension at a time, instead of relying on the resultant across all dimensions.

The $GS$ in NTGA2 -- see an example in Fig.\ref{fig:gap-scheme} -- is applied individually to each solution within the \emph{archive}, operating in the objective space. For a given solution, a single objective (chosen through a suitable heuristic) is considered. The $GS$ operator measures the distance to two neighboring points in the objective space: one neighbor with a lower objective value and another neighbor with a higher objective value. The greater of the two distances is identified as the ''gap'' distance for that solution. In cases where a solution lies on the ''edges'' of the PFa, such as points (see P(2,10) and P(9,3) in Fig.\ref{fig:gap-scheme}) where there is only one neighboring point, the ''gap'' distance in NTGA2 is assigned the value of \emph{infinity}, effectively favoring such solutions in the selection process.

It is important to note that the $GS$ utilizes a tournament selection rule, but instead of considering $fitness$ directly, it operates with $GapValues$. This approach increases the likelihood of selecting individuals that are situated close to the largest ''gaps'' in the $PFa$, thereby promoting a wider spread of the resulting $PFa$. For the selection of the second parent, a $random$ neighbor of the first individual is chosen. However, for individuals positioned on the ''edge'' of the $PFa$ approximation, there is a possibility that a second parent may not be selected, and the selection process will proceed in a similar manner to the first individual. This selection mechanism plays a crucial role in shaping the diversity and exploration of the $PFa$ during the optimization process in NTGA2.

\subsection{Strength Pareto evolutionary algorithm 2}
\label{sec:spea2}
The Strength Pareto Evolutionary Algorithm 2 (SPEA2) \cite{spea2} represents a classical approach to multi-objective optimization, leveraging environmental selection to enhance exploration of the $PFa$. SPEA2 employs an enhanced fitness assignment scheme that incorporates (1) the domination relation, (2) nearest neighbor density estimation to guide the search process, and (3) the archive truncation technique.

\begin{algorithm}[!h]
\caption{Pseudocode of SPEA2 \cite{spea2}
\label{alg:spea2}}
\begin{algorithmic}[1]
\STATE $Population \gets GenerateInitialPopulation()$
\STATE $Evaluate(Population)$
\STATE Create $Archive$
\FOR { $i \gets 0$ to $Generations$ }
    \STATE compute fitness of $Population$ and $Archive$ 
    \STATE $UpdateArchive(P_{current})$
        \IF { $|Archive| > Archive_{limit}$  }
            \STATE $truncate(Archive)$
        \ENDIF
    \STATE $Population \gets Binary\_Tournament\_Select(Population)$
    \STATE $Population \gets Crossover(Population)$
    \STATE $Population \gets Mutate(Population)$
\ENDFOR
\end{algorithmic}
\end{algorithm}

A general schema of SPEA2 is presented in Pseudocode \ref{alg:spea2}. First, initialization is applied (lines 1-3) to get a new population, evaluate individuals, and select non-dominated. In the SPEA2 main loop works the evaluation of individuals (line 5) and $Archive$ updating procedure. If $Archive$ exceeds the size (SPEA2 parameter), the specific truncate technique is launched (line 8). Then SPEA2 selects individuals (line 10) for recombination (lines 11-12) to get new individuals.  

\subsection{A Multiobjective Evolutionary Algorithm Based on Decomposition }
\label{sec:moead}
The Multiobjective Evolutionary Algorithm Based on Decomposition (MOEA/D) \cite{moead} is an evolutionary computation method used to tackle multi-objective optimization problems. It achieves this by decomposing the problem into multiple scalar subproblems, which are then optimized simultaneously. Various approaches can be employed for the decomposition process, such as Weighted Sum, Tchebycheff, or Boundary Intersection\cite{moead}. This technique allows MOEA/D to efficiently explore the multi-objective landscape, generating a diverse set of solutions that represent the trade-offs between conflicting objectives effectively.

\begin{algorithm}[!h]
\caption{Pseudocode of MOEA/D \cite{moead} with Weighted Sum approach
\label{alg:moead}}
\begin{algorithmic}[1]

\STATE $Archive \gets \emptyset$
\STATE $\lambda \gets$ ConstructWeightVectors($N$)
\STATE $B \gets$ BuildNeighborhood($\lambda$, $T$)
\STATE $Pop \gets GenerateInitialPopulation(N)$

\WHILE { $StoppingCriteria$ }
    \FOR { $i \gets 0$ to $N$ }
        \STATE $k, l \gets$ randomly select (k,l) indexes from $B(i)$
        \STATE $Ind_y \gets Crossover(Pop_k, Pop_l)$
        \STATE $Ind_y \gets Mutate(Ind_y)$
        \FOR { $j \in B(i)$ }
            \IF{$F(Ind_y|\lambda_j) < F(Pop_j|\lambda_j)$}
                \STATE $Pop_j \gets Ind_y$
            \ENDIF
        \ENDFOR
        \STATE $UpdateArchive(Ind_y)$
    \ENDFOR
\ENDWHILE
\end{algorithmic}
\end{algorithm}

MOEA/D using the basic Weighted Sum approach is presented in the Pseudocode \ref{alg:moead}. It starts with constructing $N$ weight vectors ($\lambda$) (one for each subproblem), where each vector represents a convex combination of different objectives. A single weight vector $\lambda = (\lambda_1, ..., \lambda_M)$ satisfies the constraint $\sum_{m=1}^{M}{\lambda_m}=1$, where $M$ is the number of objectives and $\lambda_m \geq 0$ is the weight for the $m$'th objective. The weight vectors are uniformly distributed, and for multi-objective instances, the Das-and-Dennis \cite{ddas} algorithm is employed for their generation. 
The neighborhood in MOEA/D is established in line 3, where $B(i)$ comprises the indexes of the $T$ closest vectors (calculated using \emph{Euclidean} distance) to $\lambda_i$, including the vector itself. In line 4, the population is randomly generated, and its size matches the number of subproblems ($N$).
The main loop proceeds until the stopping criteria are met (e.g. the number of generations). For each subproblem, the new solution $Ind_y$ is constructed (see lines 7-9 in Fig.\ref{alg:moead}) considering two neighboring solutions ($Pop_k$ and $Pop_l$). In lines 10-14, an update of neighboring solutions takes place. Each neighboring subproblem $j$ is tested; if the $Ind_y$ performs better than $Pop_j$, it takes its place. In line 15, the archive is updated based on the new solution, $Ind_y$ is added if no other solution dominates it, and all solutions dominated by $Ind_y$ are removed.

\subsection{Alternative exploration vs exploitation balancing}
\label{sec:exp}

\todo{The Monte Carlo Elites \cite{sfikas2021monte} addresses a fundamental challenge in evolutionary search methods, emphasizing the delicate balance between exploring the search space and exploiting highly fit regions. Specifically, the study extends the MAP-Elites algorithm, a quality-diversity search approach, by framing the selection of parents as a multi-armed bandit problem. This involves employing variations of the Upper-Confidence Bound (UCB) to choose parents from under-explored yet potentially rewarding areas in the search space.
The original MAP-Elites (Multi-dimensional Archive of Phenotypic Elites), an evolutionary algorithm initiates by partitioning the phenotypic space into niches, each representing distinct regions. The algorithm initializes a population randomly placed across these niches, evaluates their performance based on the objective function(s), and retains the best-performing individuals within each niche as elites. Through the reproduction process, elites coming from each niche are treated equally, ignoring their fitness or exploitation.
By applying UCB, particularly the variant derived from Monte Carlo Tree Search (UCB1 applied to trees), the paper aims to enhance the discovery of new regions and improve the overall quality of the algorithm's archive. The selection of parents is based on an indirect measure of quality, namely the survival rate of a parent's offspring. Unlike traditional methods that uniformly select individuals among retained elites, this approach introduces a focus on parent selection as a means of navigating the exploration-exploitation dilemma in quality-diversity.
The study investigates how the UCB formula, which comprises both exploitation and exploration components, influences the performance of MAP-Elites. Notably, the exploitation component is not directly tied to fitness but instead rewards individuals based on the survival of their offspring, mirroring the curiosity score in the original application. The research leverages insights from well-established node selection approaches in tree search to inform parent selection strategies in quality diversity.
This application of UCB enables a balance between exploration and exploitation in the phenotypic space. This however, requires defining criteria for partitioning solutions in the behavior/phenotypic space and determining the "success/win" measure in the UCB formula. Utilizing the offspring survival rate as a measure is a suitable choice, although in the case of multi-objective (MO) problems, determining the "success of offspring" is ambiguous. Offspring success in MO scenarios may involve being non-dominated, fully dominated, partially dominated, dominating one parent, both parents, or none.}

\todo{Another approach to dynamic and adaptive balancing of the exploration and exploitation is Adaptive Operator Selection (AOS), as exemplified by Sun et al.'s work \cite{sun2020adaptive} on Adaptive operator selection based on dynamic Thompson sampling for MOEA/D.
The efficacy of an Evolutionary Algorithm's search behavior is notably contingent on its reproduction operators, with some operators favoring exploration by effectively navigating unknown regions, while others concentrate on exploitation by refining currently superior regions. Achieving an optimal balance between these contrasting objectives has long been a central theme in the pursuit of an efficient and effective evolutionary search process. Adaptive Operator Selection (AOS) emerges as a promising paradigm aimed at autonomously choosing the most suitable reproduction operator based on the latest search dynamics. 
Typically, an AOS paradigm involves two key steps: credit assignment, which rewards an operator based on its recent performance, and decision-making, which selects the appropriate operator for the next stage according to accumulated awards. The fundamental challenge underlying AOS is the exploration versus exploitation (EvE) dilemma. To tackle this dilemma, Li et al. (\cite{li2013adaptive}) reformulated the AOS problem into a Multi-Armed Bandits (MAB) problem and applied the classic UCB algorithm to implement an AOS paradigm in Evolutionary Multi-Objective Optimization. Addressing the difficulty of credit assignment in multi-objective optimization, MOEA/D serves as a baseline, decomposing the original MOP into several single-objective subproblems to facilitate fitness evaluation. However, this is limited to the selection of mutation operators, where some operators are more random, and others rely on existing solutions. While such an approach proves effective in continuous spaces, substantial randomness in combinatorial problems may disrupt existing solutions.}

\subsection{The impact of multi-objectivity}
\label{sec:manyobj}

\todo{The prevalent approach for addressing MOP involves reliance on the dominance relation in the PFa, exemplified by algorithms like NSGA-II. However, as highlighted in \cite{fu2021adaptive}, the effectiveness of this approach significantly diminishes with an increasing number of objectives. Consequently, MOEA/D and those based on based on preference information have been proposed, focusing on specific directions in the PFa.
As discussed in \cite{allmendinger2022if}, combinatorial cases, where the number of optimal solutions exponentially grows with the number of objectives. Study conducted on multi-objective NK-landscapes with 2 to 20 objectives. Authors observed that less than 5\% of solutions are Pareto optimal for bi-objective problems (m=2), reaching approximately 50\% for m=7 objectives. Remarkably, for m=20 objectives, over 99\% of solutions are Pareto optimal for all instances considered. This indicates that the PFa not only dramatically expands with an increasing number of objectives but also constitutes a significant portion of all discovered solutions. Consequently, filtering mechanisms for the PFa based on criteria other than the dominance relation become imperative. This not only influences the algorithm's effectiveness but also its efficiency, as iterating through all PF solutions becomes computationally expensive.
Recommendations for dominance-based MOEAs, such as NSGA-II, include addressing the diminishing discriminative power of the dominance relation with an increase in objectives, which negatively impacts selection pressure and Pareto front representation quality. For decomposition/scalarization-based MOEAs, like MOEA/D, recommendations involve adjusting population size to accommodate the growing number of Pareto optimal solutions and addressing increased complexity due to the number of weight vectors. The optimal increase in weight vectors remains unclear, raising questions about the algorithm's complexity and efficiency concerning the number of objectives.}
\\
\\
\textbf{Summary}: in most of the above metaheuristics, the archive is used as a write-only collection area for non-dominated solutions. The exception is the NTGA2 method, which actively uses archive solutions to determine empty areas and influence the search direction. This is a similar idea to the population crowding distance introduced in the NSGA-II but applied on a larger scale. \todo{The essence of employing this mechanism and not relying solely on the dominance relation has been highlighted in studies concerning the impact of many-objectiveness.} It provides promising results (according to the author's results), and an interesting research direction. However, such ''forced'' steering of the algorithm has disadvantages - localized directions may have empty areas for a reason, in which case the algorithm is artificially deceived into a ''dead-end'' (i.e. constraints may reduce the number of solutions in such area). Identifying such a situation and ''withdrawing'' from further exploration of such areas would allow the algorithm to focus on more promising directions. Such observations allow us to define a new B-NTGA method that implements $balancing$ mechanism and budget search to solve considered optimization problems more effectively. \todo{As written above, existing mechanisms for balancing exploration and exploitation focus on defining dedicated metrics for space partitioning or developing a set of operators, in both cases relying on domain knowledge. Focus on the selection mechanism leverages the universal element of the method.}
\section{Problem Definition}
\label{sec:problem-deifnition}

To verify the effectiveness of the proposed B-NTGA method, we selected two practical multi- and many-objective problems: TTP)\cite{TTP} and MS-RCPSP \cite{benchmar2015}\cite{NTGA}\cite{NTGA2}. Both are NP-hard, combinatorial, and the final solution must satisfy constraints in the large landscape. Moreover, problems have conflicting objectives -- in TTP value of knapsack vs travel duration, in MS-RCPSP cost vs time of schedule -- that make those problems near real-world applications. Both problems are supported by benchmark datasets (MS-RCPSP\cite{ref:imopse} and TTP\cite{ttp-inst}) and are commonly used in literature. In this section, MS-RCPSP and TTP are defined as multi-objective optimization problems.
Moreover, the Quality Measures (QM) should be used to investigate and compare multi-objective optimization results. The QM used in this article are given in Sec.\ref{sub:metics}.  

\subsection{Multi-Skill Resource-Constrained Project Scheduling Problem}

The MS-RCPSP is a combinatorial NP-hard problem, specifically within the domain of scheduling problems. The objective is to find a \emph{feasible} schedule, meaning a solution that satisfies all constraints. This involves assigning resources to tasks and arranging them on a timeline. Feasibility of schedule $PS$ relies on meeting a set of constraints defined as follows.

Each resource is intrinsically linked to its corresponding salary, adhering to the constraint defined in Eq.\ref{eq:cost_constr}, which ensures that no salary value can assume a negative value. The subsequent aspect dictates that a resource must be associated with a non-empty set of skills, as each resource (and task) possesses connections to specific skill sets.
\begin{equation}\label{eq:cost_constr}
\forall_{r \in R} r_{salary} \geq 0 , \forall_{r \in R} S^r \neq \emptyset
\end{equation}

where $S_r$ is a set of skills possessed by the resource $r$.

Each task's duration and finish time is not negative -- according to the Eq. \ref{eq:finish_times}.

\begin{equation}\label{eq:finish_times}
\forall_{t \in T} F_t \geq 0 ; \forall_{t \in T} d_t \geq 0
\end{equation}

where $F_t$ denotes the finish time, and $d_t$ represents the duration of task $t$.
Eq.\ref{eq:prec_constr} introduces constraints associated with tasks, specifically the precedence relation among tasks. According to this constraint, work on a task can be started after all its predecessors have been completed. In other words, the tasks must follow a predefined order to maintain the correct sequence.

\begin{equation}\label{eq:prec_constr}
\forall_{t \in T, p \in t_p} F_p \leq F_t-d_t
\end{equation}
where $t_p$ represents the set encompassing all predecessors of task $t$.
The Eq.\ref{eq:skill_constr} introduces the skill extension aspect of MS-RCPSP. As per this extension, when a resource is allocated to a particular task, it is imperative that the resource possesses the skill of the required type, at a proficiency level that meets or exceeds the stipulated skill requirement. In essence, a resource must possess the necessary skills to undertake the assigned task successfully.
\begin{equation}\label{eq:skill_constr}
\forall_{t \in T^r} \; \exists_{s_r \in S^r} \; h_{s_t} = h_{s_r} \wedge l_{s_t} \leq l_{s_r}
\end{equation}

where $T^r$ is a set of tasks assigned to a resource $r$, $s_t$ is the skill required by the task $t$, $S^r$ is the set of skills possessed by the resource $r$, $h$ and $l$ are the type and level of the skill respectively.

The following constraint is defined as assigning at most one resource to any task at any given timestamp -- (see Eq.\ref{eq:once_in_time}).

\begin{equation}\label{eq:once_in_time}
\forall_{r \in R} \forall_{t \in \tau} \sum_{i=1}^n U_{i,r}^t \leq 1
\end{equation}

where $\tau$ is the time domain, $n$ represents the total number of tasks, and $U_{i,r}^t$ is a binary variable, equal to 1 if resource $r$ is assigned to task $i$ at time $t$.
The final constraint (see Eq. \ref{eq:each_task}) ensures that all tasks must be finished by ensuring that all tasks have a resource assigned at some timeslot.

\begin{equation}\label{eq:each_task}
\forall_{i \in T} \exists_{t \in \tau, r \in R}   U_{i,r}^t = 1
\end{equation}

where $\tau$ and $U_{i,r}^t$ are defined as in Eq. \ref{eq:once_in_time}.
\\
\\
The MS-RCPSP problem originally was defined as a single objective problem \cite{benchmar2015}, then extended to a bi-objective problem \cite{NTGA} and finally as a many-objective optimization problem with 5 objectives - see \cite{NTGA2}\cite{benchmark2022}. 
Two most used in literature objectives -- schedule \emph{Duration} and \emph{Cost}. Additional MS-RCPSP objective aims to describe a specific schedule aspect: \emph{Average Cash Flow}, \emph{Skill Overuse}, and the \emph{Average Use of Resources}. The MS-RSPSP optimization objectives are defined below.
\\
The {\bf Makespan} $\mathbf{f_{\tau}(PS)}$ of the project schedule $PS$ is given as Eq.\ref{eq:duration}.

\begin{equation}\label{eq:duration}
f_{\tau}(PS) = \max_{t \in T}{t_{finish}}
\end{equation}

where $T$ is a set of all tasks, $t_{finish}$ is the finish time of the task $t$.
The {\bf Cost} of the schedule is $\mathbf{f_{C}(PS)}$ defined as Eq.\ref{eq:cost}. 

\begin{equation}\label{eq:cost}
f_{C}(PS) = \sum_{i=1}^n{R_i^{salary} * T_i^{duration}}
\end{equation}

where $n$ is the number of all task-resource assignments, $R_i^{salary}$ is the salary of a resource of the i'th assignment, $T_i^{duration}$ is the duration of the task of the i'th assignment.

Skill Overuse aims to minimize the difference between the skill level of a resource and the required skill. {\bf Skill Overuse} $\mathbf{f_{S}(PS)}$ -- see Eq.\ref{eq:skill_overuse}) -- ensure that the resources assigned to the task are not overqualified, what could be essential in the practical applications.

\begin{equation}\label{eq:skill_overuse}
f_{S}(PS) = \sum_{i=1}^n{R_s^l - T_s^l}
\end{equation}

where $n$ is the number of task-resource assignments, $R_s^l$ is the skill level of a resource $R$, and $T_s^l$ is the skill level required for the task $T$.

Some resources are assigned to the project -- must receive a salary, even if they are not assigned to any task. The {\bf Average Use of Resources} (see Eq. \ref{eq:average_use_of_resources}) objective gives the distribution of tasks and ensures the efficient use of resources. It aims to minimize the deviation of the number of task-resource assignments.

\begin{equation}\label{eq:average_use_of_resources}
f_{R}(PS) = \frac{1}{r}\sum_{i = 1}^{r}{(R_i^n - R_{avg}^n)}
\end{equation}

where $r$ is the number of resources, $R_i^n$ is the number of tasks assigned to the i'th resource, $R_{avg}^n$ is the expected average number of assignments.

The {\bf Average Cash Flow} $\mathbf{f_{F}(PS)}$ (see Eq. \ref{eq:cashflow}) aims the deviation of costs over the entire duration of the project and allows for more effective budget management.

\begin{equation}\label{eq:cashflow}
f_{F}(PS) = \frac{1}{f_{\tau}(PS)}\sum_{t = 1}^{f_{\tau}}{(C_t - C_{avg})}
\end{equation}

where $C_t$ is the cost of the project in a single time slot $t$,  $f_{\tau}(PS)$ is the makespan of the project, $C_{avg}$ is the average cost of the project in a time unit and can be defined by the Eq.\ref{eq:avg_cost}.

\begin{equation}\label{eq:avg_cost}
C_{avg} = \frac{C}{f_{\tau}}
\end{equation}

where $C$ and $f_{\tau}$ are the total $cost$ and $makespan$ of the project respectively.

The multi-objective MS-RCPSP optimization uses 2 objectives ($\mathbf{f_{\tau}(PS)}$ and $\mathbf{f_{C}(PS)}$). To make the MS-RCPSP problem more useful in practice, in many-objective MS-RCPSP \cite{benchmark2022} optimization, all 5 objectives are minimized as Eq.\ref{eq:obj}.

\begin{equation}\label{eq:obj}
\min{f(PS)} = \min{[f_{\tau}(PS), f_{C}(PS), f_{F}(PS), f_{S}(PS), f_{R}(PS)]}
\end{equation}
\\
MS--RCPSP comprises two interconnected sub-problems: task sequencing, which involves placing tasks on a timeline, and resource assignments. However, not all solutions can be regarded as \emph{feasible} schedules, as certain constraints might remain unsatisfied. In the existing literature, various methodologies have been proposed to address this issue. These methods include penalizing such solutions using penalty functions, eliminating them altogether, or reconstructing/repairing them to create valid schedules. In our research, all investigated methods adopt the utilization of the \emph{Schedule Builder} based on greedy algorithm \cite{NTGA2} to obtain \emph{feasible} schedules for MS-RCPSP. This approach aids in generating schedules that fulfill the required constraints and result in viable solutions for the problem at hand.

\subsubsection{Schedule Builder} \label{sec:greedy}

Each investigated method, to get \emph{feasible} schedule uses a greedy--based algorithm -- see pseudocode \ref{alg:greedy}. The method first processes the tasks and the precedence constraints (as the predecessors). Then, other tasks are processed. The main principle is that each task is assigned at the earliest possible time it can be started. Namely, it is when all the predecessors of the tasks are finished, and its assigned resource finished its previous task assignment.

\begin{algorithm}[!h]
\caption{Schedule Builder\label{alg:greedy} for MS-RCPSP \cite{NTGA2}}
\begin{algorithmic}[2]
\FOR {task $t$}
\IF {$hasSuccesors(t)$}
\STATE $predEnd = maxFinish(t.predecessors)$
\STATE $resEnd = t.getResource().getFinish()$
\STATE $t.start = max(predEnd, resEnd)$
\ENDIF
\ENDFOR

\FOR {task t}
\IF {$!hasSuccesors(t)$}
\STATE $predEnd = maxFinish(t.predecessors)$
\STATE $resEnd = t.getResource().getFinish()$
\STATE $t.start = max(predEnd, resEnd)$
\ENDIF
\ENDFOR
\end{algorithmic}
\end{algorithm}

As it is primarily designed for duration optimization, making it particularly effective for that specific objective, its effectiveness may vary when applied to other objectives. Despite this limitation, the utilization of the \emph{ Schedule Builder} represents a well-considered compromise during the analysis of results. It is important to acknowledge that all investigated methods in this research adopt this same \emph{Schedule Builder} to construct \emph{feasible} schedules. This standardization ensures consistency in the evaluation of the methods and allows for a fair comparison of their performance in the context of the MS-RCPSP.

\subsection{Travelling Thief Problem}

The Travelling Thief Problem (TTP) \cite{TTP} combines two classical problems: Traveling Salesman Problem (TSP) and Knapsack Problem (KNP). The problem comprises a set of cities, along with their coordinates, and a set of items. Each item is described by weight, a profit, and is associated with a given city. The goal is to find a route that visits all the cities and picks up items from the cities. The Eq.\ref{eq:ttp} defines the primary goal of TTP.

\begin{equation}\label{eq:ttp}
\min{f(\pi, z)} = \min f_{\tau}(\pi, z), \max f_{P}(z)
\end{equation}

where $\pi$ and $z$ are the permutations of cities visited and the picking plan. The objective $f_{\tau}$ is to minimize the traveling plan. The $f_{P}$ objective is to maximize the profit from the picked items. The relation between those problems is that picking items decreases travel speed.

\begin{equation}\label{eq:travelling_time}
f_{\tau}(\pi, z) = \sum_{i=1}^{n-1}\frac{d_{\pi_{i},\pi_{i+1}}}{v(w(\pi_{i}))} + \frac{d_{\pi_{n},\pi_{1}}}{v(w(\pi_{n}))}
\end{equation}

where $d_{\pi_{i},\pi_{i+1}}$ is the distance between two consecutive cities, $n$ is a number of cities , $v(w(\pi_{i}))$ is the velocity in city $\pi_i$, which depends on weight $w$. The second factor of the equation ensures that the route is completed by returning to the first city.

The entire traveling time (see Eq.\ref{eq:travelling_time}) is modified as items are picked, and velocity decreases according to Eq.\ref{eq:velocity}.

\begin{equation}\label{eq:velocity}
v(w)=v_{max}-\frac{W_c}{W}(v_{max} - v_{min})
\end{equation}

where $W_c$ and $W$ are the current and maximum allowed weights. The model defines the speed: maximum $v_{max}$ and minimum $v_{min}$ speed depending on $W$. The weight $w$ is defined as the sum of all weights of items picked up so far.

\begin{equation}\label{eq:profit}
f_{P}(z) = \sum_{j=1}^m z_j z_j^{profit}
\end{equation}

where $m$ is the number of items, $z_j$ is equal to 1 if the $j$'th item has been picked, 0 otherwise. $z_j^{profit}$ is the profit of the $j$'th item.

The Eq.\ref{eq:profit} describes the profit as the second TTP objective.

For the picking plan to be feasible, the constraint KNP (see Eq.\ref{eq:weight_constraint}) must be satisfied.

\begin{equation}\label{eq:weight_constraint}
\sum_{j=1}^m z_j z_j^{weight} \leq W
\end{equation}

where $m$ is the number of items, $z_j$ is defined as in Eq.\ref{eq:profit}, and $z_j^{weight}$ is the weight of $j$'th item. The equation ensures that the weight of picked items does not exceed the limit $W$ of the maximum allowed knapsack weight.
\\
\\
Some additional assumptions should be made to investigate methods for solving the multi-objective optimization problem. \emph{PerfectPoint} is defined as the point with the best possible values for all objectives. For MS-RCPSP, the best makespan is defined as the duration of the shortest task multiplied by the number of tasks divided by the number of resources. The best cost value is the cost of the schedule where all tasks are assigned to the cheapest resource. The best values of skill overuse, average use of resources, and average cash flow are equal to 0. For TTP, the reference value of minimum time is defined as the total length of the minimum spanning tree divided by the maximum speed. The reference value of maximum profit is achieved by a brute-force algorithm starting from the items with the highest profit/weight ratio. \emph{NadirPoint} is defined as a point with the worst possible values for all objectives. For the MS-RCPSP, the worst value of makespan is the total sum of all tasks' duration. The worst cost value is the cost of schedule where all tasks are performed by the most expensive resource. Worst skill overuse is achieved by assigning tasks to the resources with the highest level of required skill. The worst average use of resources is the maximum makespan multiplied by the number of resources - 1 and divided by the number of resources. The worst value of the average cash flow is the same as the maximum cost. For the TTP, the worst time is twice the value of the minimum time. The reference value of minimum profit is equal to 0. 
\section{Balanced Non-dominated Tournament Genetic Algorithm}
\label{sec:method}

This section contains the main motivations of the proposed Balanced Non-dominated Tournament Genetic Algorithm (B-NTGA) method, motivations, its assumptions, and essential components. 

\subsection{Motivation}
\label{sub:motivation}

The proposed B-NTGA is inspired by NTGA2\cite{NTGA2} (see Sec.\ref{sub:ntga2}) research. Analysis of the NTGA2 and its results revealed potential areas of improvement by incorporating ideas from the literature. The main novelty of the NTGA2 (see sec.\ref{sub:ntga2}) is the \emph{Gap Selection} ($GS$) operator. Its goal is to promote exploration of the least explored regions of PFa. The original assumptions suggest that wider ‘‘gap'' in the PFa implies missing solutions, thus the higher exploration potential. While this is partially true, authors do not consider the possibility of natural ‘‘gaps'', where constraints may cause that in some PFa regions, feasible solutions may not exist (e.g. in MS-RCPSP). Thus, it can lead to significant computational budget waste and increase the risk of the method being stuck in the local optima. The authors also try to force population diversity by the clones elimination method. It can help the population leave occasional local optima, but it might become a waste of computational power when it happens too frequently. 

Mainly, the B-NTGA uses a \emph{balancing} mechanism to deal with computational budget waste and allows B-NTGA to be more effective than NTGA2. Moreover, B-NTGA, compared to NTGA2, uses only one selection method- redesigned $GS$ with the balance mechanism. Also, B-NTGA does not use the clone elimination procedure. Thus, the B-NTGA schema is simpler than NTGA2 and uses fewer parameters -- \todo{more detailed comparison of B-NTGA and NTGA2 can be found in sec.\ref{sub:ntga2vsbntga}}. The B-NTGA method has been subjected to experimental tests described in the section \ref{sec:experiments}, which resulted in the method presented below. 

\subsection{Balanced Non-dominated Tournament Genetic Algorithm}
\label{sub:bgtga}

The B-NTGA is the many-objective optimization metaheuristic based on a genetic algorithm. Thus, in the proposed B-NTGA, standard genetic operators are used, but the problem-specific operators could be more effective. However, the main attention has been dedicated to the ‘‘gap'' calculation and selection mechanism. Additionally, balancing the archive exploration and reducing the oversampling of non-promising areas allowed the removal of the clone elimination and second selection operator, which appeared to be redundant in B-NTGA. 

\begin{algorithm}[!h]
  \caption{Balanced Non-dominated Tournament Genetic Algorithm (B-NTGA)
   \label{alg:main}}
  \begin{algorithmic}[1]
    \STATE \textbf{Params:} $PopSize,\:Generations,\:P_{x},\:P_{m},\:TourSize$
    \STATE $archive \gets \emptyset$ 
    \STATE $P \gets InitPopulation(PopSize)$
    \STATE $UpdateArchive(P)$
    \FOR{$g \gets 0$ to $Generations$}
      \STATE $GapValues = CalculateGapValues(archive)$
      \FOR{$i \gets 0$ to $PopSize / 2$}
        \STATE $Parents \gets GapTourSelection(archive,\:GapValues,\:TourSize)$
        \STATE $IncrementSelectionCount(Parents)$
        \STATE $Children \gets Crossover(Parents,\:P_{x})$
        \STATE $P[2i, 2i+1] \gets Mutate(Children,\:P_{m})$
      \ENDFOR
      \STATE $UpdateArchive(P)$
    \ENDFOR
    \STATE \textbf{Return:} $archive$
  \end{algorithmic}
\end{algorithm}

The B-NTGA uses random initialization of population $P$ of $PopSize$ size (line 3 in Pseudocode \ref{alg:main}). Then $UpdateArchive$ (line 4) function takes a vector of individuals and updates $archive$ with non-dominated (by $P$ or $archive$) individuals and removes individuals from $achive$ which got dominated. In line 5, the main loop begins until the specified number of generations passes. In line 6, the $GapValues$ vector is calculated (details are given in sec.\ref{sec:gapValue}). Lines 7 to 12 present how a new population is generated using selection, crossover, and mutation operators. In contrast to NTGA2's double selection, a single selection operator is used. Individuals for the new population are created by pairs (thus a number of required loops in line 7 is $PopSize/2$) - two $Parents$ are selected from the $archive$ using the $GapTourSelection$ method described below. Their genotypes are copied and crossed over using a standard uniform crossover operator with $P_{x}$ probability. $IncrementSelectionCount$ (line 9) is the novel B-NTGA step, where the individuals selected from the \emph{archive} have their ‘‘selection counters" increased by one. This counter is used by the $GapTourSelection$ described in sec.\ref{sec:gapValue}. In line 10, newly generated genotypes are mutated using standard \emph{random bit} mutation with per gene $P_{m}$ mutation probability. In line 11, generated and evaluated individuals replace two individuals in the population. Note that population is not used anywhere inside the selection loop. Then $archive$ is updated (line 13). Concluding the algorithm, complete $archive$ consisting of all found non-dominated solutions is returned as the results.

\subsubsection{Balanced Gap Selection mechanism}\label{sec:gapValue} \
The $GapTourSelection$ uses a standard tournament mechanism (size of $TourSize$) to find a pair of parent individuals, where $GapValues$ are used instead of fitness. The $GapValue$ is a property calculated for each individual stored in the \emph{archive}. The function starts with a randomly selected dimension used for \emph{archive} sorting (ascending) by the corresponding objective value. Next, for each element in the sorted \emph{archive}, the distance (in the given dimension) to both neighbors (one with a worse objective value and one with better) is calculated, and a larger distance is selected (see Fig.\ref{fig:gap-scheme} in Sec.\ref{sub:ntga2}). In the work\cite{NTGA2} the ‘‘Gap Selection'' is introduced and proposed calculating distances for each dimension and selecting the maximum among them. The ‘‘edge'' individuals were considered to have \emph{infinite} distance and their $GapValue$ was set to \emph{infinity} value to match it. After a series of experiments (e.g. $Round-Robin$ scheme has been tested), there were no significant differences in the calculating distance for each dimension over a random one. However, it could reduce the proposed method's stability and research reproducibility, but a random dimension is preferred due to its efficiency and simplicity.

\begin{figure}[!ht]
\centering
\begin{adjustbox}{width=1.0\textwidth,center=\textwidth}
\includegraphics[width=1.0\textwidth]{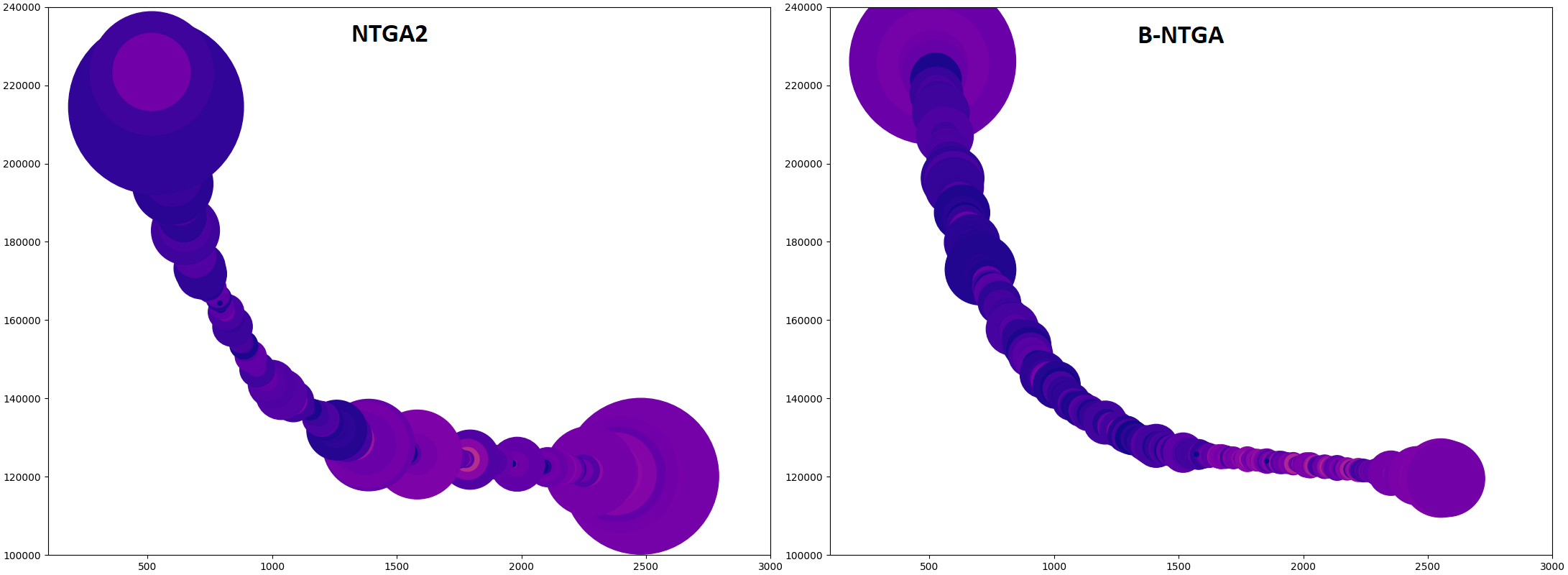}
\end{adjustbox}
\caption{An illustration of oversampling in NTGA2 vs. B-NTGA [used: MS-RCPSP instance 200\_10\_84\_9]}
\label{fig:gap-sel-comp}
\end{figure}

An illustration in Fig.\ref{fig:gap-sel-comp} is given to demonstrate the oversampling situation and motivate the proposed B-NTGA. The figure shows how NTGA2 / B-NTGA deals with PFa exploration/exploitation and how some PFa regions are frequently visited. Each sphere represents the single individual (solution) stored in the $archive$ (after 2000 generations), where the sphere's radius reflects its selection count. High-radius spheres in the middle area of the NTGA2's PFa mark the larger ‘‘gaps''. Thus, nearby individuals are frequently selected. As B-NTGA considers selection count, it decreases the selection pressure for highly-exploited individuals.

Results of experiments showed that another crucial aspect of selection is handling the ''edge'' points. Using $infinity$ value benefits ''stretching'' the PFa, but it also introduces a risk of oversampling those points (see Fig.\ref{fig:gap-sel-comp}), even if there is a low probability for improvement. The figure shows that the ‘‘edges'' of PFa are frequently selected because of \emph{infinity} value. To make the \emph{Gap Selection} (using $GapValues$) more balanced in B-NTGA some modifications were examined. Firstly, to use a distance to the \emph{PerfectPoint} for the point in the ‘‘lower edge'' and the ‘‘upper edge'' - to use its distance to the \emph{NadirPoint} point instead of $infinity$. Another examined modification is to set a hard limit on how many times an individual can be selected. However, none of the examined modifications improved the overall results.

The main contribution of this paper applies to the $GapValue$ calculation itself. The original method (NTGA2) narrows to the distance calculation solely. Inspired by the \emph{Niching}\cite{ref:niching} algorithms and \emph{Monte Carlo Elites}\cite{sfikas2021monte} driven by \emph{Upper Confidence Bound} (UCB)\cite{upper} formula, attempts have been made to enhance the balance between exploration and exploitation of points in the archive. Unlike the other density-only methods, B-NTGA considers the level of exploitation of PFa areas, which is tracked by the ‘‘selection counter" stored per individual in the archive. Figure \ref{fig:gap-sel-comp} illustrates the comparison of the results between the original $GapValue$ calculation in NTGA2 and the balanced approach in B-NTGA. The B-NTGA $PFa$ is more balanced, and there is no such significant domination of some $PFa$ regions in the selection operator.

B-NTGA uses a modified \emph{Monte Carlo Elites} selection -- see Eq.\ref{eq:ucb}\cite{sfikas2021monte} to calculate the chance to select $i$ individual ($U_i$) as follows.
\begin{equation}
\label{eq:ucb}
U_i = \frac{w_i}{ns_i} + \lambda \sqrt{\frac{\ln{(NS)}}{ns_i}}
\end{equation}
where $w_i$ is the number of offspring survivals for individual $i$, $ns_i$ is the current number of selections for individuals $i$, $NS$ is the total number of all selections so far, and $\lambda$ is a constant set to $1/\sqrt{2}$. The outcome $U_i$ value is assigned to individual $i$, reflecting its chance of being selected. The first part of the equation represents the exploitation of the solution, while the second part represents the exploration. 

Thus Eq.\ref{eq:ucb} can be adapted effectively to the \emph{Gap Selection} operator in different ways; a straightforward approach would result in the equation where the $Gap_i$ replaces $w_i$. Where $Gap_i$ is the value of ‘‘gap'' distance for the individual $i$ in the random dimension calculated by minimal $Euclidean$ distance (see an example in Fig. \ref{fig:gap-scheme} in sec.\ref{sub:ntga2}). However, the approach becomes very fragile to the objective domains as values can differ in scale.

\begin{equation}
\label{eq:gapc}
ComplexGapValue_i = \frac{Gap_i}{ns_i} + \lambda \sqrt{\frac{\ln{(NS * MaxSpan)}}{ns_i}}
\end{equation}

A simple approach to normalization would be to divide the $Gap_i$ by the $MaxSpan$ equal to the highest value found in the objective space - lowest value found in the objective space; or to multiply $MaxSpan$ by $NS$. The Eq. \ref{eq:gapc} presents the modified UCB equation where $Gap_i$, calculated according to the above description, creates the exploitation factor. Its value is divided by the $ns_i$ number of individual $i$ selected from the archive. Thus, at the beginning to avoid dividing by 0, this value is set to 1. The second part of the equation represents the ''exploitation'' factor. $NS$ is the total number of selections at that iteration, equal to $PopSize * g$, where $g$ is the current generation number. One additional parameter $\lambda$ is introduced, balancing the exploration vs exploitation ratio. In literature\cite{sfikas2021monte}, it is common to set this value to either $\sqrt{2}$ or $1/\sqrt{2}$ as the starting point and then tune it according to the needs. If the $\lambda$ value equals 0, it effectively ignores the second factor and results in a simplified version equal to $Gap_i / ns_i$. 

All of these variants were evaluated experimentally. However, a simplified version appeared to be as competitive and dropped the $\lambda$ parameter; therefore, Eq.\ref{eq:gaps} is the suggested approach to the implementation when parameter tuning is a time-consuming process. It presents the final $GapValue$ version, which has been experimentally verified, selected, and used throughout further experiments.

\begin{equation}
\label{eq:gaps}
GapValue_i = \frac{Gap_i}{ns_i}
\end{equation}

To summarise, the \textbf{balanced} Gap tournament selection in B-NTGA has been effectively simplified compared to the original $Gap$ selection operator originally used in NTGA2 \cite{NTGA2}. The operator receives the \emph{archive} sorted during the $CalculateGapValues$ step described above, with a $GapValues$ vector in corresponding order. The first parent selection remains unchanged and follows the standard $tournament$ selection, where $GapValue$ is used as the fitness. The second parent is a random neighboring individual in the sorted \emph{archive}. If the point lies on the ‘‘edge'' of the $PFa$ (either the best or worst solution in terms of a considered objective), the only viable neighbor is selected, instead of using the recursive $tournament$ selection (which took place in the NTGA2).

\subsection{B-NTGA vs. NTGA2 -- comparision}
\label{sub:ntga2vsbntga}

There is a list of essential differences (see Tab.\ref{tab:bntga-ntga2}) between the elements of B-NTGA and its inspiration - NTGA2 method. The applied GAP selection has been improved with a balancing mechanism, which increases the diversity of explored solutions. According to the results obtained during the experiments, it bested the clone elimination mechanism (used in NTGA2), which rejected individuals repeated in the population and generated new ones. However, in most cases, the appearance of clones meant getting stuck in the local optimum and resulted in generating clones many times, which reduced the method's efficiency. The improved GAP selection mechanism incorporates the adaptive mechanism which rebalances the importance of certain PFa areas throughout the evolution process. The key element is the actively updated counter associated with each solution in \emph{archive}. As the new method is \emph{archive} selection only and does not store dominated individuals in the population. Its presence is not mandatory -- further research on this outcome is being conducted in parallel.

\begin{table}[!ht]
\caption{B-NTGA vs. NTGA2 -- the main differences \label{tab:bntga-ntga2}}
\begin{adjustbox}{width=1.0\textwidth,center=\textwidth}
\begin{tabular}{|l|c||c|}
\hline
\emph{feature} & B-NTGA & NTGA2 \cite{ntg2}   \\ \hline
selection type & balanced (tournament) GAP & (1) tournament GAP, (2) tournament  \\ \hline
parameters & 4+1 & 6+1 \\ \hline
diversification strategy & balanced selection & 2 selections + clone elimination mechanism \\ \hline
exploration throughout evolution & balanced, adaptive GAP & static GAP \\ \hline
selection source & \emph{archive} & (1) population, (2) \emph{archive}  \\ \hline
\emph{archive} usage & active & write-only + selection   \\ \hline
population role in searching & auxiliary & base \\ \hline
\end{tabular}
\end{adjustbox}
\end{table}

Moreover, the proposed B-NTGA uses a single selection type, instead of two (NTGA2). It reduces two parameters, i.e. $gsGen$, $TourSize$ (from 2 sizes to 1 size parameter), see Tab. \ref{tab:parameters}, which makes the B-NTGA method easier to tune and control. Moreover, the indirect parameter ($+1$) defines the computational budget given as the number of generations.
\section{Experiments and Results}
\label{sec:experiments}

The main goal of developed experiments is to verify the effectiveness and efficiency of the proposed B-NTGA solving two NP-hard multi- and many-objective optimization problems (TTP and MS-RCPSP) compared to state-of-the-art methods. To investigate B-NTGA, the following {\bf Research Questions} have been developed:

 \begin{itemize}
  \item {\bf \emph{RQ0.}} How to tune the methods used in experiments? How to measure the results of multi- and many-objective optimization?
  \item {\bf \emph{RQ1.}} How ‘‘good" does B-NTGA achieve the PFa compared to reference methods solving multi-objective TTP?
  \item {\bf \emph{RQ2.}} Is B-NTGA useful in solving multi-objective MS-RCPSP comparing results to state-of-the-art methods?
    \item {\bf \emph{RQ3.}} Is B-NTGA effective is solving many--objective MS-RCPSP with 5 objectives?
    \item {\bf \emph{RQ4.}} How B-NTGA is more effective than other approaches solving multi- and many-objective optimisation?
\end{itemize}

Mainly, this section contains research details and answers to the above Research Questions. To answer $RQ0$ the multi-objective quality measures (QMs) have been presented in sec.\ref{sub:metics}, and the tuning procedure and configurations for used methods are given sec.\ref{sub:parameters} and the used instances are given in sec.\ref{sub:instances}. The used experimental procedure is given in sec.\ref{sub:procedure}. The main section presents investigation results for proposed B-NTGA compared to state-of-the-art methods for two benchmark multi-objective problems: TTP (see sec.\ref{sub:results-2obj-TTP}) to answer $RQ1$ and MS-RCPSP (see sec.\ref{sub:results-2obj-MS-RCPSP}) to response for $RQ2$. Results for many-objective problems are presented in sec. \ref{sub:results-5obj-MS-RCPSP}) are connected to $RQ3$ question. The last section (see sec.\ref{sub:discussion}) contains an answer to $RQ4$ as a summary and discussion of gained results.
 
\subsection{Metrics and Quality Measures}
\label{sub:metics}

In multi- or many--objective optimization, the main goal is to find a set of non-dominated solutions. If a solution has a value of at least one objective better and values of no objectives worse than another, then the given solution dominates another (domination relation). The true Pareto Front (TPF) could be defined as a set of all non-dominated solutions. However, in practical real-world problems (like TTP and MS-RCPSP) TPF usually is unknown, and all results obtained are \emph{approximations} of the PF (PFa).

This subsection describes all the Quality Measures (QMs) used to compare experimental results. Hence the nature of multi- and many-objective optimization, the output of each method is given as PFa, which is built of a set of non-dominated (ND) points found by the investigated method. Point is dominated if any other point has at least one better (lower, considering both evaluated problems) objective value and no worse objective value. All selected QMs are calculated based on the returned PFa.

The commonly used multi-objective metric in the literature \cite{Survey} is Hypervolume (HV) but its exponential cost and higher number of objectives make it problematic to use in the many-objective area. As an alternative, Inverted Generative Distance\cite{Survey} (IGD) is used as the primary metric as it captures both convergence and diversity. It is an average distance from each TPF point to the closest point in PF as presented in Eq.\ref{eq:igd}, where $d_i$ is the \emph{Euclidean} distance for the $i$-th point. As objectives vary in scale, using absolute values for IGD calculation might favor certain objectives. For that reason, points in PF are normalized beforehand. Every point is normalized in a standard manner using minimum and maximum values from the TPF.\\

\begin{equation}
\label{eq:igd}
IGD(PF, TPF) = \frac{\sqrt{\sum_{i=1}^{|TPF|}d_i^2}}{|TPF|}
\end{equation}

$Purity$ \cite{Survey} has been selected for the secondary QMs. It is defined as in Eq.\ref{eq:purity}, where $ND$ is the number of solutions (aggregated from all runs) not dominated by the ‘‘True Pareto Front approximation" (TPFa), where TPFa is constructed by merging all PFa (results of all investigated methods) and removing dominated solutions. $Purity$ calculated for a single method returns the value from 0 to 1 and could be interpreted as the size part of TPFa that the given method resulted in. However, the same points (solutions) can be found by different methods. Therefore summed up $Purity$ for all investigated methods could exceed the value of 1.

\begin{equation}
\label{eq:purity}
Purity(PF, TPF) = \frac{|ND(PF, TPF)|}{|TPF|}
\end{equation}

These two $QMs$ give information about PFa generated by a given method -- how it dominates others ($Purity$) and how ‘‘close" (to $PP$) and ‘‘diversified" is ($IGD$) in solution space.

\subsection{Experimental Procedure}
\label{sub:procedure}

Some methods (e.g. MOEA/D) use reference points in calculations (objectives normalization). Not required by others (like NTGA2 or B-NTGA) as they treat objectives separately. They are calculated as proposed in the work \cite{NTGA2}. 

The single result of each run is a set of non-dominated solutions found for a given instance. Solutions are being saved using absolute coordinates in the objective space. Different objectives may vary in domain space, and even the bigger disproportions occur across different instances, those coordinates should be normalized first. Reference points (\emph{Pefect Point}  and \emph{Nadir Point} -- defined in sec.\ref{sec:problem-deifnition}) are used for the standard normalization process, which projects coordinates to [0, 1] vectors. It prevents favoring algorithms from performing better in ‘‘higher-value" objectives.

IGD is a relative metric, it uses \emph{True Pareto Front} (TPF) as a reference point. As the real TPF is unknown, it is constructed using results generated by all runs of all methods. IGD is calculated for each of the 30 runs, then averaged and put into the results table. Purity is the number of points from all 30 runs merged, which are not dominated by the TPF, constructed as described above.
To verify the statistical significance of presented results Wilcoxon signed-rank test is used.

\subsection{Instances}
\label{sub:instances}

In experiments, published iMOPSE \cite{ref:imopse} dataset (presented in \cite{benchmar2015} or \cite{benchmark2022}) is used, which contains 36 MS-RCPSP instances created using real-world scheduling problems. All instances have varying tasks, resources, and skills to define a range of problems.
The dataset contains two main parts: 100 and 200 tasks. The number of resources varies between 5 and 20 for the first part and 10 and 40 for the second part. The number of skills varies between 9 and 15.

The benchmark dataset for TTP presented in \cite{blank2017solving} has been selected for TTP. Instances based on the \emph{eil51} instance are commonly used (e.g. \cite{blank2017solving}\cite{NTGA2}). Selected TTP 12 instances differ in varying items per city (between 50 and 500) and could be divided into three groups. They are: 
(1) with a strong correlation between item weights and profits.
(2) instances where weights and profits are completely uncorrelated.
and (3) instances where there is no correlation, but all the items have similar weights.

The tested method has been evaluated on a full 36-instance set of MS-RCPSP and 12 instances of TTP \cite{ttp-inst}. Due to the non-deterministic nature of evolutionary computation, all runs have been repeated 30 times, and results have been averaged. 

\subsection{Configurations}
\label{sub:parameters}

Before the main experiments were conducted, all investigated methods had to be tuned first. The 5-Level Taguchi \cite{nair1992taguchi} Parameter Design procedure has been used to conduct a systematic parameter tuning. A set of experiment configurations was generated using an orthogonal matrix, and each configuration was repeated 10 times. This procedure was further repeated for a subset of test functions. The parameters with the highest Signal-to-Noise change were fine-tuned first based on the average results. All the parameters have been processed in that manner subsequently in the descending order of Signal-to-Noise change. The number of generations was adapted to always match the constant number of maximum births/fitness evaluations. For the MS-RCPSP it was set to 50.000 (2obj) and 250.000 (5obj). For the TTP it was set to 250.000 (2obj). Moreover, for MS-RCPSP problem(s) (2-obj and 5-obj versions) U-NSGA-III has been used, and for 2-objective TTP NSGA-II is used as a reference method. Additionally, configurations of NTGA2, U-NSGA-III, and $\theta$-DEA for MS-RCPSP are based on work \cite{NTGA2}.

\begin{table}
\caption{The best found configurations for investigated methods\label{tab:parameters}}
\begin{adjustbox}{width=1.0\textwidth,center=\textwidth}
\begin{tabular}{ll|c|c|c|c|c|c|c|c}
~ & ~ & PopSize &  $P_{m}$ & $P_{x}$ & TourSize & $\theta$ & gsGen & NhSize & PartNr  \\ \hline
MS-RCPSP (2obj) & B-NTGA & 50 & 0.01 & 0.9 & 40 & ~ & ~ & ~ &   \\ 
~ & NTGA2 & 50 & 0.01 & 0.6 & 6 / 20 & ~ & 50 & ~ &   \\ 
~ & $\theta$-DEA & 100 & 0.005 & 0.6 & 2 & 0.5 & ~ & ~ &   \\ 
~ & U-NSGA-III & 100 & 0.005 & 0.9 & 2 & ~ & ~ & ~ &   \\ 
~ & MOEA/D & (50) & 0.015 & 0.2 & ~ & ~ & ~ & 6 & 50  \\ 
~ & SPEA2 & 200 & 0.015 & 0.99 & ~ & ~ & ~ & ~ &   \\ 
~ & NSGA-II & 300 & 0.015 & 0.99 & 2 & ~ & ~ & ~ &   \\ 
\hline
MS-RCPSP (5obj) & B-NTGA & 50 & 0.01 & 0.9 & 40 & ~ & ~ & ~ & \\ 
~ & NTGA2 & 50 & 0.01 & 0.6 & 6 / 20 & ~ & 50 & ~ &  \\ 
~ & $\theta$-DEA & 500 & 0.005 & 0.6 & 2 & 0.5 & ~ & ~ & \\ 
~ & U-NSGA-III & 200 & 0.005 & 0.9 & 2 & ~ & ~ & ~ & \\ 
~ & MOEA/D & (330) & 0.02 & 0.4 & ~ & ~ & ~ & 2 & 7 \\ 
~ & SPEA2 & 400 & 0.015 & 0.9 & ~ & ~ & ~ & ~ & \\ 
~ & NSGA-II & 400 & 0.015 & 0.9 & 2 & ~ & ~ & ~ & \\ 
\hline
TTP & B-NTGA & 50 & 0.7 / 0.7 & 0.5 / 0.5 & 40 & ~ & ~ & ~ & \\ 
~ & NTGA2 & 50 & 0.9 / 0.9 & 0.3 / 0.3 & 6 / 20 & ~ & 50 & ~ &   \\ 
~ & $\theta$-DEA & 100 & 0.9 / 0.9 & 0.3 / 0.3 & ~ & 0.1 & ~ & ~ &  \\ 
~ & MOEA/D & (100) & 0.4 / 0.3 & 0.5 / 1.0 & ~ & ~ & ~ & 3 & 100 \\ 
~ & SPEA2 & 100 & 0.4 / 0.3 & 0.1 / 0.8 & ~ & ~ & ~ & ~ &  \\ 
~ & NSGA-II & 300 & 0.4 / 0.7 & 0.9 / 0.3 & 2 & ~ & ~ & ~ & \\ 
\end{tabular}
\end{adjustbox}
\end{table}

All investigated methods use the same solution landscape. As representation (genotype) for both problems, we use a discrete encoding. For MS-RCPSP it is an association-type encoding - genotype has a length equal to the number of tasks to be scheduled, and each position is an index of the valid resource. Finally, to get \emph{feasible} schedule greedy schedule Builder (see sec.\ref{sec:greedy}) is used. For the TTP, it is a permutation-type + binary encoding - the path is encoded by ordered indices of cities, while knapsack consists of either \emph{1s} if the item is selected, or \emph{0s} if not.

Table \ref{tab:parameters} presents the best-found configurations used in the main experiments. Population Size ($PopSize$) is the constant number of individuals in a generation. For the MOEA/D, instead of explicitly set, it is based on the number of decomposition vectors constructed using \cite{ddas} algorithm for the given Partitions Number ($PartNr$). Mutation Probability ($P_{m}$) is a probability on a scale $[0, 1]$. For the MS-RCPSP it is a probability of a single gene's random mutation. For the TTP it is split into two values. First is the probability of the random path segment being reversed. Second, is the probability of a random item decision change ($bit flip$). Crossover Probability ($P_{x}$) is the probability of two individuals crossing over. All implemented methods use the same Uniform Crossover operator for the MS-RCPSP and a combination of OX (route) with SX (knapsack) for the TTP. Tournament Size ($TourSize$) is the number is individuals considered whenever the tournament selection operator is used. Based on the original NTGA2 implementation, 2 values were found to be used, first for the Standard Tournament and second for the GAP Tournament Selection. Theta is the $\theta$-DEA specific parameter. A number of generations ($gsGen$) is the selection switch parameter used by the NTGA2. The neighborhood Size ($NhSize$) is the number of adjacent decomposition vectors considered by MOEA/D when solutions are compared.


\subsection{Results for multi-objective TTP}
\label{sub:results-2obj-TTP}

The complete comparison of results for all tested methods (NTGA2, $\theta$-DEA, MOEA/D, SPEA2, NSGA-II ad B-NTGA) to answer $RQ1$ has been developed series of experiments. Results are given in Table \ref{tab:ttp-igd} (IGD) and \ref{tab:ttp-nd} ($Purity$). Each table contains the average and standard deviation of QMs for every method. At the bottom of each table can be found the complete average of the method and Wilcoxon signed-rank test results on whether B-NTGA has improved the results. Additionally, on the right column of each table, the technical column has been added to validate B-NTGA's performance against all other methods combined (selected best result other than B-NTGA in each row). 

\begin{table}[!h]
\caption{Results for multi-objective TTP (2obj) using IGD $[10^{-3}]$
\label{tab:ttp-igd}
}
\begin{adjustbox}{width=1.1\textwidth,center=\textwidth}
\begin{tabular}{l|rr|rr|rr|rr|rr|rr||rr}
 & \multicolumn{2}{c|}{\textbf{B-NTGA}}                          & \multicolumn{2}{c|}{\textbf{NTGA2}}                                   & \multicolumn{2}{c|}{\textbf{$\theta$-DEA}}                         & \multicolumn{2}{c|}{\textbf{MOEA/D}}                         & \multicolumn{2}{c|}{\textbf{SPEA2}}                         & \multicolumn{2}{c||}{\textbf{NSGA-II}}                        & \multicolumn{2}{c}{{[}$non$ B-NTGA{]}}                              \\
instance                                                      & \multicolumn{1}{c}{\textbf{avg}} & \multicolumn{1}{c|}{std} & \multicolumn{1}{c}{\textbf{avg}}      & \multicolumn{1}{c|}{std}      & \multicolumn{1}{c}{\textbf{avg}} & \multicolumn{1}{c|}{std} & \multicolumn{1}{c}{\textbf{avg}} & \multicolumn{1}{c|}{std} & \multicolumn{1}{c}{\textbf{avg}} & \multicolumn{1}{c|}{std} & \multicolumn{1}{c}{\textbf{avg}} & \multicolumn{1}{c||}{std} & \multicolumn{1}{c}{\textbf{avg}} & \multicolumn{1}{c}{std} \\
\hline
eil51\_n50\_bounded-strongly-corr\_01   & 7.02          & 2.78 & \textbf{6.86}  & 2.69 & 94.53  & 16.16 & 15.66 & 4.80  & 12.68 & 4.38  & 17.76 & 6.74  & \textbf{6.86}  & 2.69 \\
eil51\_n50\_uncorr\_01                  & 8.91          & 3.54 & \textbf{7.57}  & 2.42 & 98.68  & 12.35 & 16.32 & 6.75  & 14.15 & 4.70  & 18.45 & 6.45  & \textbf{7.57}  & 2.42 \\
eil51\_n50\_uncorr-similar-weights\_01  & 30.03         & 9.76 & \textbf{20.35} & 7.70 & 285.50 & 63.31 & 53.92 & 18.36 & 98.01 & 38.96 & 45.32 & 14.67 & \textbf{20.35} & 7.70 \\
eil51\_n150\_bounded-strongly-corr\_01  & \textbf{2.02} & 0.84 & 2.85           & 0.89 & 29.15  & 4.78  & 4.25  & 1.34  & 7.08  & 2.23  & 6.72  & 1.99  & 2.85           & 0.89 \\
eil51\_n150\_uncorr\_01                 & \textbf{3.54} & 1.08 & 4.25           & 1.77 & 57.25  & 7.87  & 6.22  & 2.58  & 11.12 & 2.55  & 12.97 & 3.29  & 4.25           & 1.77 \\
eil51\_n150\_uncorr-similar-weights\_01 & \textbf{5.79} & 2.35 & 7.96           & 2.26 & 98.46  & 14.72 & 13.87 & 5.20  & 13.85 & 5.18  & 14.99 & 5.33  & 7.96           & 2.26 \\
eil51\_n250\_bounded-strongly-corr\_01  & \textbf{1.55} & 0.58 & 2.94           & 0.79 & 24.54  & 3.11  & 3.89  & 1.10  & 7.76  & 1.69  & 7.60  & 1.68  & 2.94           & 0.79 \\
eil51\_n250\_uncorr\_01                 & \textbf{1.35} & 0.53 & 2.08           & 0.75 & 31.56  & 3.34  & 2.55  & 0.94  & 8.08  & 2.12  & 8.94  & 1.77  & 2.08           & 0.75 \\
eil51\_n250\_uncorr-similar-weights\_01 & \textbf{2.48} & 1.32 & 3.84           & 1.62 & 43.37  & 6.19  & 6.04  & 2.35  & 11.77 & 2.82  & 11.42 & 2.48  & 3.84           & 1.62 \\
eil51\_n500\_bounded-strongly-corr\_01  & \textbf{1.30} & 0.49 & 3.15           & 0.49 & 21.79  & 2.23  & 4.08  & 0.68  & 9.80  & 1.28  & 11.17 & 1.58  & 3.15           & 0.49 \\
eil51\_n500\_uncorr\_01                 & \textbf{0.91} & 0.44 & 2.76           & 0.55 & 25.92  & 2.63  & 2.37  & 0.52  & 9.85  & 1.21  & 10.34 & 1.17  & 2.37           & 0.52 \\
eil51\_n500\_uncorr-similar-weights\_01 & \textbf{1.31} & 0.57 & 2.95           & 0.95 & 28.79  & 3.81  & 3.55  & 1.20  & 9.27  & 1.84  & 8.06  & 1.72  & 2.95           & 0.95 \\
\hline
avg                                     & \textbf{5.52} & 2.02 & 5.63           & 1.91 & 69.96  & 11.71 & 11.06 & 3.82  & 17.79 & 5.75  & 14.48 & 4.07  & 5.60           & 1.90 \\
stat                                                          & \textbf{}                        &                         & \multicolumn{2}{l|}{\textbf{$\approx$(P \textless 0.20)}} & \multicolumn{2}{l|}{\textbf{+ (P \textless 0.001)}}         & \multicolumn{2}{l|}{\textbf{+ (P \textless 0.001)}}         & \multicolumn{2}{l|}{\textbf{+ (P \textless 0.001)}}         & \multicolumn{2}{l||}{\textbf{+ (P \textless 0.001)}}         & \multicolumn{2}{l}{\textbf{$\approx$(P \textless 0.20)}}
\end{tabular}
\end{adjustbox}
\end{table}

Results presented in Table \ref{tab:ttp-igd} show that B-NTGA outperforms all methods for most instances. The exception exists in three (\_n50) instances, where NTGA2 gives better solutions. The $Purity$ measure (see Tab.\ref{tab:ttp-nd}) confirms better NTGA2 results for three instances. Moreover, results presented in Tab.\ref{tab:ttp-nd} confirm that both methods (NTGA2 and B-NTGA) completely dominate other state-of-the-art methods, which additionally confirms values of Wilcoxon signed-rank test results. Additionally, statistical analysis confirmed that B-NTGA is better than NTGA2 if three (\_n50) instances are not considered. 

\begin{table}[!h]
\caption{Results for multi-objective TTP (2obj) using \emph{Purity}
\label{tab:ttp-nd}
}
\begin{adjustbox}{width=1.1\textwidth,center=\textwidth}
\begin{tabular}{l|rr|rr|rr|rr|rr|rr||rr}
  & \multicolumn{2}{c|}{\textbf{B-NTGA}}                          & \multicolumn{2}{c|}{\textbf{NTGA2}}                               & \multicolumn{2}{c|}{\textbf{$\theta$-DEA}}                         & \multicolumn{2}{c|}{\textbf{MOEA/D}}                         & \multicolumn{2}{c|}{\textbf{SPEA2}}                         & \multicolumn{2}{c||}{\textbf{NSGA-II}}                        & \multicolumn{2}{c}{{[}$non$ B-NTGA{]}}                          \\
instance                                & \multicolumn{1}{c}{\textbf{avg}} & \multicolumn{1}{c|}{std} & \multicolumn{1}{c}{\textbf{avg}}    & \multicolumn{1}{c|}{std}    & \multicolumn{1}{c}{\textbf{avg}} & \multicolumn{1}{c|}{std} & \multicolumn{1}{c}{\textbf{avg}} & \multicolumn{1}{c|}{std} & \multicolumn{1}{c}{\textbf{avg}} & \multicolumn{1}{c|}{std} & \multicolumn{1}{c}{\textbf{avg}} & \multicolumn{1}{c||}{std} & \multicolumn{1}{c}{\textbf{avg}}    & \multicolumn{1}{c}{std}    \\
\hline
eil51\_n50\_bounded-strongly-corr\_01   & \textbf{0.431}          & 0.055                   & \textbf{0.431}          & 0.067                   & 0.000                   & 0.000                   & 0.010                   & 0.002                   & 0.118                   & 0.021                   & 0.010                   & 0.002                   & \textbf{0.431}          & 0.067                   \\
eil51\_n50\_uncorr\_01                  & 0.235                   & 0.027                   & \textbf{0.518}          & 0.073                   & 0.000                   & 0.000                   & 0.141                   & 0.023                   & 0.000                   & 0.000                   & 0.012                   & 0.002                   & \textbf{0.518}          & 0.073                   \\
eil51\_n50\_uncorr-similar-weights\_01  & 0.286                   & 0.027                   & \textbf{0.486}          & 0.061                   & 0.000                   & 0.000                   & 0.114                   & 0.012                   & 0.000                   & 0.000                   & 0.000                   & 0.000                   & \textbf{0.486}          & 0.061                   \\
eil51\_n150\_bounded-strongly-corr\_01  & \textbf{0.675}          & 0.087                   & 0.279                   & 0.036                   & 0.000                   & 0.000                   & 0.038                   & 0.007                   & 0.002                   & 0.000                   & 0.004                   & 0.001                   & 0.279                   & 0.036                   \\
eil51\_n150\_uncorr\_01                 & 0.267                   & 0.047                   & \textbf{0.457}          & 0.062                   & 0.000                   & 0.000                   & 0.276                   & 0.028                   & 0.000                   & 0.000                   & 0.000                   & 0.000                   & \textbf{0.457}          & 0.062                   \\
eil51\_n150\_uncorr-similar-weights\_01 & \textbf{0.816}          & 0.146                   & 0.000                   & 0.000                   & 0.000                   & 0.000                   & 0.098                   & 0.014                   & 0.074                   & 0.013                   & 0.000                   & 0.000                   & 0.098                   & 0.014                   \\
eil51\_n250\_bounded-strongly-corr\_01  & \textbf{0.985}          & 0.106                   & 0.000                   & 0.000                   & 0.000                   & 0.000                   & 0.011                   & 0.001                   & 0.000                   & 0.000                   & 0.000                   & 0.000                   & 0.011                   & 0.001                   \\
eil51\_n250\_uncorr\_01                 & \textbf{0.967}          & 0.095                   & 0.001                   & 0.000                   & 0.000                   & 0.000                   & 0.000                   & 0.000                   & 0.000                   & 0.000                   & 0.000                   & 0.000                   & 0.001                   & 0.000                   \\
eil51\_n250\_uncorr-similar-weights\_01 & \textbf{0.860}          & 0.108                   & 0.000                   & 0.000                   & 0.000                   & 0.000                   & 0.024                   & 0.003                   & 0.000                   & 0.000                   & 0.000                   & 0.000                   & 0.024                   & 0.003                   \\
eil51\_n500\_bounded-strongly-corr\_01  & \textbf{0.848}          & 0.107                   & 0.000                   & 0.000                   & 0.000                   & 0.000                   & 0.031                   & 0.005                   & 0.000                   & 0.000                   & 0.120                   & 0.022                   & 0.120                   & 0.022                   \\
eil51\_n500\_uncorr\_01                 & \textbf{0.705}          & 0.111                   & 0.000                   & 0.000                   & 0.000                   & 0.000                   & 0.295                   & 0.037                   & 0.000                   & 0.000                   & 0.000                   & 0.000                   & 0.295                   & 0.037                   \\
eil51\_n500\_uncorr-similar-weights\_01 & \textbf{0.846}          & 0.089                   & 0.000                   & 0.000                   & 0.000                   & 0.000                   & 0.140                   & 0.019                   & 0.000                   & 0.000                   & 0.000                   & 0.000                   & 0.140                   & 0.019                   \\
\hline
avg                                     & \textbf{0.660}          & 0.084                   & 0.181                   & 0.025                   & 0.000                   & 0.000                   & 0.098                   & 0.013                   & 0.016                   & 0.003                   & 0.012                   & 0.002                   & 0.238                   & 0.033                  \\
stat                                    & \textbf{}                        &                         & \multicolumn{2}{l|}{\textbf{$\approx$ (P \textless 0.02)}} & \multicolumn{2}{l|}{\textbf{+ (P \textless 0.001)}}         & \multicolumn{2}{l|}{\textbf{+ (P \textless 0.001)}}         & \multicolumn{2}{l|}{\textbf{+ (P \textless 0.001)}}         & \multicolumn{2}{l||}{\textbf{+ (P \textless 0.001)}}         & \multicolumn{2}{l}{\textbf{$\approx$ (P \textless 0.02)}}
\end{tabular}
\end{adjustbox}
\end{table}

\begin{figure}[!h]
\centering
\includegraphics[width=0.95\textwidth]{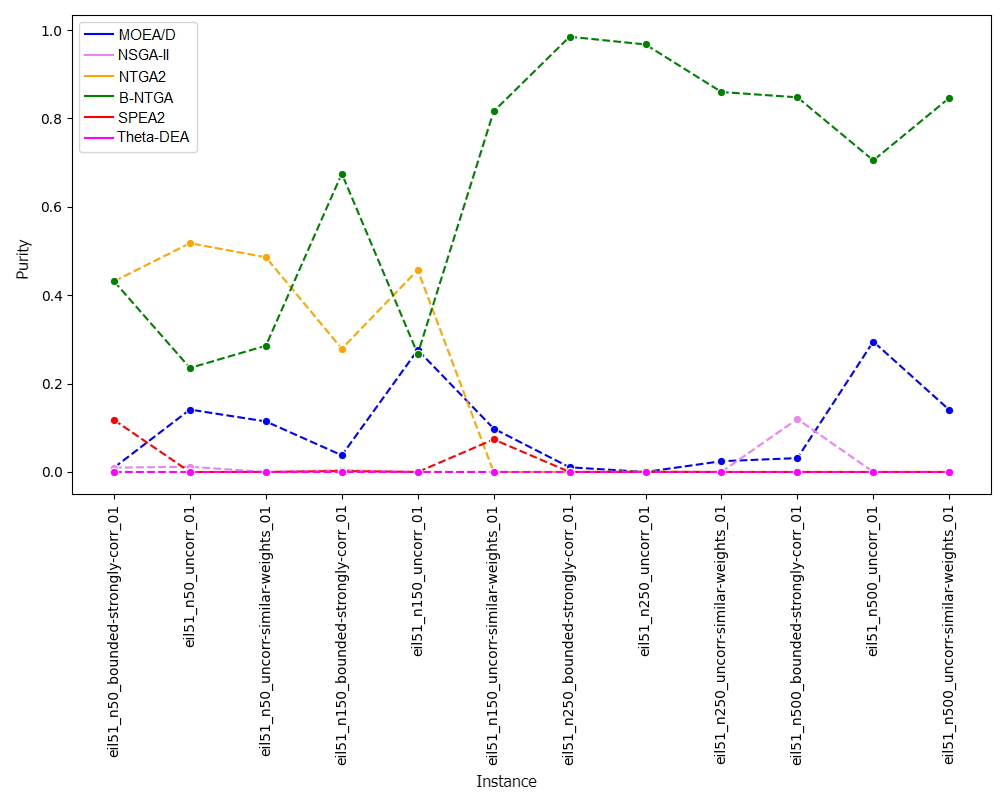}
\caption{Summary of results for multi-objective TTP using Purity}
\label{fig:ttp2d-all-MND-TPFS}
\end{figure}

Fig. \ref{fig:ttp2d-all-MND-TPFS} graphically presents the relations of how examined methods build the final PFa for each TTP instance. It's worth mentioning that for most instances (9/12) the B-NTGA gives works out the larger (67\% or more) portion of PFa.

\begin{figure}[!h]
\centering
\includegraphics[width=1.0\textwidth]{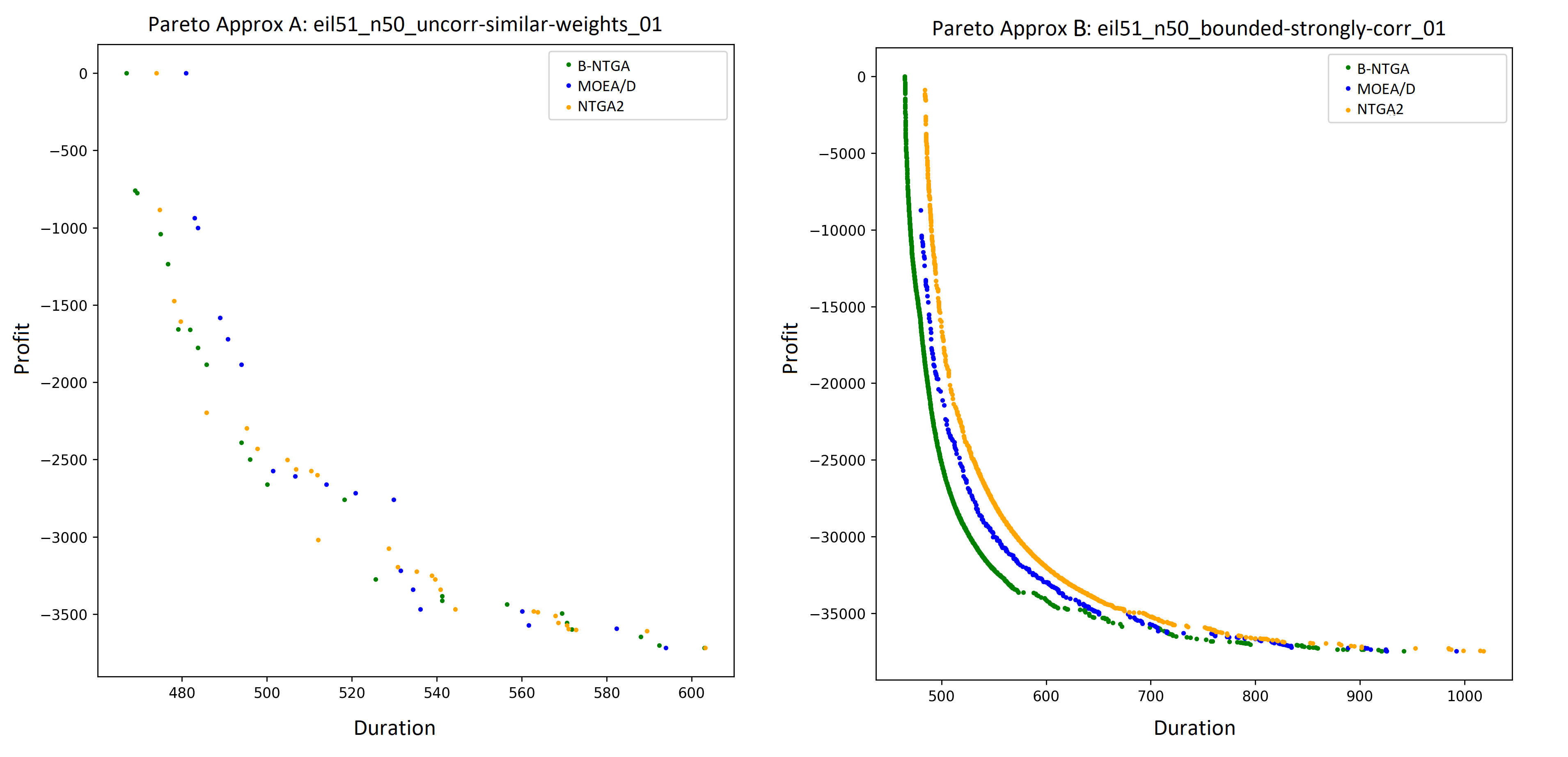}
\caption{Two examples of PFa for TTP instances -- with sparse (a) and dense (b) PFa.}
\label{fig:ttp2d-PFa}
\end{figure}

In TTP problem, three instances are problematic for B-NTGA ($IGD$ and $Purity$ context), where NTGA2 outperforms it. In these three instances, the resulting PFa is sparse. Fig.\ref{fig:ttp2d-PFa} shows examples of two PFa types of results. Instance (a) shows sparse PFa, where B-NTGA is less efficient. Another instance (b) presents a situation where B-NTGA outperforms other methods.


\subsection{Results for multi-objective MS-RCPSP}
\label{sub:results-2obj-MS-RCPSP}

Experiments that use multi-objective MS-RCPSP optimize only two objectives (schedule cost and duration) -- in contrast to 5-objective optimization -- make easier analysis and interpretation of gained results. Table \ref{tab:2-msrpsp-igd} results for all examined methods are presented using IGD measure. In most cases (34/36) B-NTGA outperforms other methods, which answers $RQ2$. The results are statistically significant by using Wilcoxon signed-rank test.

\begin{table}[!h]
\caption{Results for multi-objective MS-RCPSP (2obj) using IGD $[10^{-3}]$
\label{tab:2-msrpsp-igd}}
\begin{adjustbox}{width=1.1\textwidth,center=\textwidth}
\begin{tabular}{l|rr|rr|rr|rr|rr|rr|rr||rr}
 & \multicolumn{2}{c|}{\textbf{B-NTGA}}                          & \multicolumn{2}{c|}{\textbf{NTGA2}}                     & \multicolumn{2}{c|}{\textbf{$\theta$-DEA}}                         & \multicolumn{2}{c|}{\textbf{U-NSGA-III}}                    & \multicolumn{2}{c|}{\textbf{MOEA/D}}                         & \multicolumn{2}{c|}{\textbf{SPEA2}}                         & \multicolumn{2}{c||}{\textbf{NSGA-II}}                        & \multicolumn{2}{c}{{[}$non$ B-NTGA{]}}                    \\
instance                                                          & \multicolumn{1}{c}{\textbf{avg}} & \multicolumn{1}{c|}{std} & \multicolumn{1}{c}{\textbf{avg}} & \multicolumn{1}{c|}{std} & \multicolumn{1}{c}{\textbf{avg}} & \multicolumn{1}{c|}{std} & \multicolumn{1}{c}{\textbf{avg}} & \multicolumn{1}{c|}{std} & \multicolumn{1}{c}{\textbf{avg}} & \multicolumn{1}{c|}{std} & \multicolumn{1}{c}{\textbf{avg}} & \multicolumn{1}{c|}{std} & \multicolumn{1}{c}{\textbf{avg}} & \multicolumn{1}{c||}{std} & \multicolumn{1}{c}{\textbf{avg}} & \multicolumn{1}{c}{std} \\
\hline
100\_10\_26\_15     & \textbf{1.38} & 0.74 & 1.69  & 0.65 & 9.23  & 2.06 & 6.90  & 1.76 & 1.53          & 0.73 & 3.06          & 1.32 & 2.87  & 1.33 & 1.53          & 0.73  \\
100\_10\_27\_9\_D2  & \textbf{0.79} & 0.38 & 1.56  & 0.76 & 12.96 & 1.67 & 11.90 & 1.40 & 1.85          & 0.56 & 10.51         & 1.27 & 9.32  & 1.79 & 1.56          & 0.76  \\
100\_10\_47\_9      & \textbf{0.37} & 0.18 & 0.75  & 0.34 & 11.28 & 1.41 & 10.99 & 1.24 & 0.57          & 0.15 & 6.49          & 0.76 & 6.50  & 0.96 & 0.57          & 0.15  \\
100\_10\_48\_15     & 1.84          & 1.12 & 2.02  & 0.63 & 6.86  & 1.78 & 4.22  & 1.65 & 2.31          & 0.67 & \textbf{1.35} & 0.25 & 1.86  & 0.94 & \textbf{1.35} & 0.25  \\
100\_10\_64\_9      & \textbf{0.83} & 0.32 & 1.52  & 0.50 & 12.28 & 1.74 & 11.49 & 1.40 & 1.06          & 0.26 & 7.46          & 1.13 & 7.39  & 1.22 & 1.06          & 0.26  \\
100\_10\_65\_15     & \textbf{1.01} & 0.46 & 1.18  & 0.47 & 6.44  & 1.77 & 5.35  & 1.57 & 1.46          & 0.37 & 3.93          & 0.81 & 4.92  & 0.73 & 1.18          & 0.47  \\
100\_20\_22\_15     & \textbf{1.72} & 0.66 & 2.07  & 0.91 & 17.48 & 1.43 & 15.33 & 2.22 & 2.65          & 0.92 & 11.78         & 2.48 & 12.42 & 2.10 & 2.07          & 0.91  \\
100\_20\_23\_9\_D1  & \textbf{1.28} & 0.48 & 2.27  & 1.06 & 18.29 & 2.28 & 16.06 & 1.79 & 3.60          & 1.29 & 12.47         & 2.43 & 13.30 & 1.53 & 2.27          & 1.06  \\
100\_20\_46\_15     & \textbf{3.00} & 0.90 & 3.39  & 0.76 & 11.62 & 2.90 & 9.02  & 2.11 & 3.54          & 0.97 & 6.17          & 2.22 & 7.29  & 2.11 & 3.39          & 0.76  \\
100\_20\_47\_9      & \textbf{1.00} & 0.19 & 2.94  & 0.73 & 17.87 & 1.42 & 16.76 & 1.24 & 2.28          & 0.91 & 14.24         & 1.20 & 15.09 & 1.08 & 2.28          & 0.91  \\
100\_20\_65\_15     & \textbf{1.61} & 0.45 & 1.84  & 0.63 & 12.48 & 3.49 & 10.41 & 2.88 & 4.64          & 1.19 & 8.97          & 2.94 & 10.92 & 2.77 & 1.84          & 0.63  \\
100\_20\_65\_9      & \textbf{0.79} & 0.27 & 5.48  & 1.28 & 18.25 & 1.71 & 16.71 & 1.38 & 3.46          & 0.75 & 12.44         & 1.20 & 12.15 & 1.48 & 3.46          & 0.75  \\
100\_5\_20\_9\_D3   & \textbf{0.25} & 0.11 & 0.49  & 0.19 & 8.96  & 1.51 & 6.33  & 1.27 & 0.42          & 0.14 & 8.39          & 0.63 & 3.70  & 1.10 & 0.42          & 0.14  \\
100\_5\_22\_15      & \textbf{0.83} & 0.13 & 1.10  & 0.47 & 1.34  & 0.84 & 1.00  & 0.34 & 1.67          & 0.50 & 2.59          & 3.38 & 1.15  & 0.15 & 1.00          & 0.34  \\
100\_5\_46\_15      & \textbf{0.87} & 0.63 & 1.31  & 0.56 & 1.61  & 0.64 & 1.63  & 0.52 & 1.20          & 0.62 & 9.09          & 3.38 & 1.09  & 0.38 & 1.09          & 0.38  \\
100\_5\_48\_9       & \textbf{0.36} & 0.32 & 0.70  & 0.42 & 2.25  & 1.59 & 2.56  & 1.52 & 0.42          & 0.31 & 6.68          & 1.71 & 0.59  & 0.18 & 0.42          & 2.25  \\
100\_5\_64\_15      & \textbf{0.58} & 0.25 & 0.97  & 0.62 & 3.12  & 1.24 & 2.84  & 1.02 & 0.76          & 0.32 & 4.20          & 2.11 & 2.00  & 0.54 & 0.76          & 0.32  \\
100\_5\_64\_9       & \textbf{0.14} & 0.07 & 0.38  & 0.25 & 10.30 & 1.34 & 9.67  & 1.41 & 0.35          & 0.08 & 9.40          & 0.38 & 5.22  & 2.34 & 0.35          & 0.08  \\
200\_10\_128\_15    & \textbf{0.84} & 0.43 & 1.82  & 0.65 & 13.01 & 1.71 & 10.85 & 1.45 & 1.30          & 0.22 & 5.96          & 1.39 & 7.56  & 1.44 & 1.30          & 0.22  \\
200\_10\_135\_9\_D6 & \textbf{2.07} & 0.42 & 2.93  & 0.48 & 14.20 & 1.64 & 13.37 & 1.87 & 3.51          & 0.64 & 9.41          & 2.09 & 11.33 & 1.74 & 2.93          & 0.48  \\
200\_10\_50\_15     & 1.67          & 0.68 & 2.69  & 0.89 & 15.22 & 1.47 & 12.69 & 1.30 & \textbf{1.60} & 0.45 & 8.86          & 1.08 & 10.60 & 1.15 & \textbf{1.60} & 0.45  \\
200\_10\_50\_9      & \textbf{0.25} & 0.10 & 4.72  & 0.58 & 13.82 & 0.86 & 13.64 & 0.89 & 1.16          & 0.32 & 10.16         & 0.89 & 11.24 & 0.67 & 1.16          & 0.32  \\
200\_10\_84\_9      & \textbf{0.24} & 0.11 & 5.52  & 0.89 & 14.86 & 0.66 & 14.52 & 0.83 & 3.50          & 1.03 & 11.83         & 0.63 & 12.13 & 0.48 & 3.50          & 1.03  \\
200\_10\_85\_15     & \textbf{0.22} & 0.09 & 1.28  & 0.49 & 12.96 & 1.06 & 13.17 & 1.02 & 0.49          & 0.10 & 9.50          & 1.02 & 10.15 & 0.94 & 0.49          & 12.96 \\
200\_20\_145\_15    & \textbf{0.86} & 0.24 & 7.65  & 1.06 & 19.83 & 1.25 & 19.64 & 1.22 & 5.58          & 0.80 & 15.74         & 0.95 & 15.47 & 1.09 & 5.58          & 0.80  \\
200\_20\_150\_9\_D5 & \textbf{4.53} & 0.96 & 10.43 & 1.78 & 33.37 & 5.01 & 30.49 & 3.51 & 19.03         & 3.41 & 43.87         & 5.73 & 36.29 & 4.00 & 10.43         & 1.78  \\
200\_20\_54\_15     & \textbf{1.32} & 0.36 & 2.08  & 0.46 & 17.17 & 1.37 & 16.11 & 1.74 & 1.71          & 0.20 & 10.19         & 1.69 & 10.12 & 1.66 & 1.71          & 0.20  \\
200\_20\_55\_9      & \textbf{1.35} & 0.64 & 8.79  & 0.79 & 20.21 & 0.95 & 20.20 & 0.91 & 6.47          & 0.56 & 15.70         & 0.84 & 16.35 & 0.57 & 6.47          & 0.56  \\
200\_20\_97\_15     & \textbf{0.68} & 0.25 & 7.08  & 1.37 & 22.96 & 1.12 & 21.94 & 1.44 & 5.24          & 1.20 & 19.23         & 1.33 & 18.69 & 1.53 & 5.24          & 1.20  \\
200\_20\_97\_9      & \textbf{1.76} & 0.43 & 5.75  & 0.65 & 17.66 & 1.23 & 17.33 & 1.22 & 4.23          & 0.33 & 12.45         & 1.23 & 12.43 & 1.00 & 4.23          & 0.33  \\
200\_40\_130\_9\_D4 & \textbf{8.58} & 1.91 & 13.49 & 2.42 & 41.73 & 5.48 & 41.50 & 7.77 & 16.82         & 1.94 & 52.30         & 6.46 & 44.82 & 4.99 & 13.49         & 2.42  \\
200\_40\_133\_15    & \textbf{2.54} & 0.32 & 7.26  & 0.74 & 21.64 & 1.35 & 20.71 & 1.78 & 5.68          & 0.75 & 17.15         & 1.78 & 16.57 & 1.30 & 5.68          & 0.75  \\
200\_40\_45\_15     & \textbf{2.10} & 0.27 & 8.04  & 1.19 & 21.62 & 1.03 & 20.94 & 1.23 & 5.48          & 0.78 & 17.49         & 1.05 & 17.77 & 1.15 & 5.48          & 0.78  \\
200\_40\_45\_9      & \textbf{1.95} & 0.45 & 10.04 & 0.99 & 23.76 & 0.97 & 22.97 & 0.77 & 8.59          & 0.84 & 20.15         & 0.97 & 20.56 & 0.85 & 8.59          & 0.84  \\
200\_40\_90\_9      & \textbf{2.87} & 0.53 & 8.69  & 0.80 & 22.22 & 1.11 & 21.56 & 1.02 & 6.01          & 0.75 & 18.18         & 1.22 & 18.07 & 1.11 & 6.01          & 0.75  \\
200\_40\_91\_15     & \textbf{3.13} & 0.42 & 6.67  & 0.90 & 19.31 & 1.28 & 18.72 & 1.23 & 5.47          & 0.57 & 15.86         & 1.30 & 14.68 & 1.15 & 5.47          & 0.57  \\
\hline
avg                 & \textbf{1.54} & 0.45 & 4.07  & 0.79 & 15.23 & 1.68 & 14.15 & 1.61 & 3.77          & 0.71 & 12.59         & 1.70 & 11.74 & 1.38 & 3.22          & 1.04 \\
stat                                                              & \textbf{}                        &                         & \multicolumn{2}{l|}{\textbf{+ (P \textless 0.001)}}         & \multicolumn{2}{l|}{\textbf{+ (P \textless 0.001)}}         & \multicolumn{2}{l|}{\textbf{+ (P \textless 0.001)}}         & \multicolumn{2}{l|}{\textbf{+ (P \textless 0.001)}}         & \multicolumn{2}{l|}{\textbf{+ (P \textless 0.001)}}         & \multicolumn{2}{l||}{\textbf{+ (P \textless 0.001)}}         & \multicolumn{2}{l}{\textbf{+ (P \textless 0.001)}}        
\end{tabular}
\end{adjustbox}
\end{table}

In Table \ref{tab:2-msrpsp-mnd} are presented results for all experimentally tested methods using the $Purity$ measure. The details show that in most cases (29/36) the B-NTGA gives better results than other methods. Interestingly, in 6 cases MOEA/D dominates B-NTGA. However, the average results of B-NTGA give the first prize, and the result is statistically significant. It is worth noticing, that on average B-NTGA gives the PFa 3x better than NTGA2 and 1.6x better than MOEA/D.

\begin{table}[!h]
\caption{Results for multi-objective MS-RCPSP (2obj) using \emph{Purity}
\label{tab:2-msrpsp-mnd}}
\begin{adjustbox}{width=1.1\textwidth,center=\textwidth}
\begin{tabular}{l|rr|rr|rr|rr|rr|rr|rr||rr}
& \multicolumn{2}{c|}{\textbf{B-NTGA}}    &
\multicolumn{2}{c|}{\textbf{NTGA2}}                            & \multicolumn{2}{c|}{\textbf{$\theta$-DEA}}                            & \multicolumn{2}{c|}{\textbf{U-NSGA-III}}                       & \multicolumn{2}{c|}{\textbf{MOEA/D}}                            & \multicolumn{2}{c|}{\textbf{SPEA2}}                            & \multicolumn{2}{c||}{\textbf{NSGA-II}}                           & \multicolumn{2}{c}{{[}$non$ B-NTGA{]}}                                       \\
instance                    & \multicolumn{1}{c}{\textbf{avg}} & \multicolumn{1}{c|}{std} & \multicolumn{1}{c}{\textbf{avg}} & \multicolumn{1}{c|}{std}    & \multicolumn{1}{c}{\textbf{avg}} & \multicolumn{1}{c|}{std}    & \multicolumn{1}{c}{\textbf{avg}} & \multicolumn{1}{c|}{std}    & \multicolumn{1}{c}{\textbf{avg}} & \multicolumn{1}{c|}{std}    & \multicolumn{1}{c}{\textbf{avg}} & \multicolumn{1}{c|}{std}    & \multicolumn{1}{c}{\textbf{avg}} & \multicolumn{1}{c||}{std}    & \multicolumn{1}{c}{\textbf{avg}} & \multicolumn{1}{c}{std}                    \\
\hline
100\_10\_26\_15     & \textbf{0.464} & 0.009 & 0.329 & 0.045 & 0.097 & 0.005 & 0.079 & 0.005 & 0.337          & 0.017 & 0.145          & 0.009 & 0.074 & 0.006 & 0.337          & 0.017 \\
100\_10\_27\_9\_D2  & \textbf{0.598} & 0.024 & 0.288 & 0.022 & 0.072 & 0.007 & 0.025 & 0.003 & 0.099          & 0.006 & 0.027          & 0.004 & 0.018 & 0.002 & 0.288          & 0.022 \\
100\_10\_47\_9      & \textbf{0.645} & 0.009 & 0.369 & 0.053 & 0.053 & 0.002 & 0.031 & 0.001 & 0.479          & 0.022 & 0.048          & 0.002 & 0.004 & 0.000 & 0.479          & 0.022 \\
100\_10\_48\_15     & 0.200          & 0.009 & 0.046 & 0.003 & 0.090 & 0.004 & 0.061 & 0.003 & 0.101          & 0.004 & \textbf{0.322} & 0.011 & 0.055 & 0.002 & \textbf{0.322} & 0.011 \\
100\_10\_64\_9      & \textbf{0.600} & 0.070 & 0.355 & 0.032 & 0.047 & 0.002 & 0.045 & 0.004 & 0.459          & 0.047 & 0.012          & 0.001 & 0.005 & 0.001 & 0.459          & 0.047 \\
100\_10\_65\_15     & \textbf{0.320} & 0.006 & 0.151 & 0.016 & 0.076 & 0.004 & 0.039 & 0.002 & 0.178          & 0.004 & 0.079          & 0.006 & 0.018 & 0.001 & 0.178          & 0.004 \\
100\_20\_22\_15     & \textbf{0.406} & 0.028 & 0.078 & 0.006 & 0.081 & 0.004 & 0.027 & 0.002 & 0.203          & 0.005 & 0.006          & 0.001 & 0.003 & 0.001 & 0.203          & 0.005 \\
100\_20\_23\_9\_D1  & \textbf{0.720} & 0.078 & 0.017 & 0.002 & 0.036 & 0.003 & 0.010 & 0.001 & 0.060          & 0.004 & 0.012          & 0.002 & 0.000 & 0.000 & 0.060          & 0.004 \\
100\_20\_46\_15     & \textbf{0.303} & 0.014 & 0.098 & 0.005 & 0.118 & 0.009 & 0.031 & 0.002 & 0.220          & 0.010 & 0.024          & 0.003 & 0.000 & 0.000 & 0.220          & 0.010 \\
100\_20\_47\_9      & \textbf{0.707} & 0.095 & 0.011 & 0.001 & 0.035 & 0.002 & 0.017 & 0.002 & 0.116          & 0.005 & 0.006          & 0.001 & 0.000 & 0.000 & 0.116          & 0.005 \\
100\_20\_65\_15     & \textbf{0.360} & 0.013 & 0.315 & 0.037 & 0.077 & 0.004 & 0.041 & 0.004 & 0.077          & 0.003 & 0.018          & 0.002 & 0.018 & 0.002 & 0.315          & 0.037 \\
100\_20\_65\_9      & \textbf{0.598} & 0.093 & 0.017 & 0.001 & 0.060 & 0.004 & 0.014 & 0.001 & 0.216          & 0.005 & 0.000          & 0.000 & 0.002 & 0.000 & 0.216          & 0.005 \\
100\_5\_20\_9\_D3   & \textbf{0.764} & 0.018 & 0.659 & 0.026 & 0.087 & 0.005 & 0.126 & 0.012 & 0.575          & 0.060 & 0.394          & 0.041 & 0.580 & 0.082 & 0.659          & 0.026 \\
100\_5\_22\_15      & \textbf{0.450} & 0.010 & 0.370 & 0.009 & 0.402 & 0.075 & 0.413 & 0.046 & 0.392          & 0.027 & 0.402          & 0.123 & 0.397 & 0.025 & 0.413          & 0.046 \\
100\_5\_46\_15      & \textbf{0.576} & 0.010 & 0.404 & 0.007 & 0.389 & 0.007 & 0.389 & 0.011 & 0.355          & 0.021 & 0.338          & 0.029 & 0.382 & 0.020 & 0.404          & 0.007 \\
100\_5\_48\_9       & \textbf{0.499} & 0.007 & 0.287 & 0.007 & 0.246 & 0.046 & 0.274 & 0.052 & 0.341          & 0.017 & 0.263          & 0.043 & 0.366 & 0.025 & 0.366          & 0.025 \\
100\_5\_64\_15      & \textbf{0.485} & 0.022 & 0.280 & 0.022 & 0.219 & 0.017 & 0.212 & 0.022 & 0.233          & 0.014 & 0.219          & 0.018 & 0.181 & 0.007 & 0.280          & 0.022 \\
100\_5\_64\_9       & \textbf{0.839} & 0.017 & 0.775 & 0.028 & 0.090 & 0.004 & 0.216 & 0.028 & 0.766          & 0.053 & 0.187          & 0.041 & 0.516 & 0.135 & 0.775          & 0.028 \\
200\_10\_128\_15    & \textbf{0.510} & 0.010 & 0.110 & 0.009 & 0.105 & 0.006 & 0.007 & 0.001 & 0.192          & 0.009 & 0.004          & 0.000 & 0.006 & 0.001 & 0.192          & 0.009 \\
200\_10\_135\_9\_D6 & \textbf{0.475} & 0.035 & 0.019 & 0.003 & 0.075 & 0.009 & 0.003 & 0.001 & 0.253          & 0.013 & 0.003          & 0.001 & 0.072 & 0.013 & 0.253          & 0.013 \\
200\_10\_50\_15     & \textbf{0.461} & 0.023 & 0.052 & 0.003 & 0.077 & 0.003 & 0.015 & 0.001 & 0.193          & 0.005 & 0.021          & 0.002 & 0.013 & 0.001 & 0.193          & 0.005 \\
200\_10\_50\_9      & \textbf{0.623} & 0.075 & 0.057 & 0.003 & 0.030 & 0.002 & 0.024 & 0.002 & 0.336          & 0.012 & 0.001          & 0.000 & 0.001 & 0.000 & 0.336          & 0.012 \\
200\_10\_84\_9      & \textbf{0.741} & 0.096 & 0.017 & 0.001 & 0.026 & 0.002 & 0.010 & 0.001 & 0.149          & 0.004 & 0.000          & 0.000 & 0.000 & 0.000 & 0.149          & 0.004 \\
200\_10\_85\_15     & \textbf{0.604} & 0.032 & 0.124 & 0.009 & 0.044 & 0.002 & 0.026 & 0.001 & 0.541          & 0.018 & 0.000          & 0.000 & 0.000 & 0.000 & 0.541          & 0.018 \\
200\_20\_145\_15    & \textbf{0.597} & 0.025 & 0.010 & 0.001 & 0.019 & 0.001 & 0.031 & 0.003 & 0.263          & 0.006 & 0.000          & 0.000 & 0.000 & 0.000 & 0.263          & 0.006 \\
200\_20\_150\_9\_D5 & \textbf{0.578} & 0.029 & 0.059 & 0.011 & 0.157 & 0.027 & 0.070 & 0.013 & 0.059          & 0.007 & 0.000          & 0.000 & 0.000 & 0.000 & 0.157          & 0.027 \\
200\_20\_54\_15     & \textbf{0.465} & 0.027 & 0.008 & 0.001 & 0.064 & 0.004 & 0.006 & 0.001 & 0.326          & 0.009 & 0.003          & 0.001 & 0.000 & 0.000 & 0.326          & 0.009 \\
200\_20\_55\_9      & 0.378          & 0.026 & 0.009 & 0.001 & 0.037 & 0.004 & 0.022 & 0.002 & \textbf{0.462} & 0.012 & 0.000          & 0.000 & 0.000 & 0.000 & \textbf{0.462} & 0.012 \\
200\_20\_97\_15     & \textbf{0.599} & 0.056 & 0.018 & 0.001 & 0.041 & 0.004 & 0.050 & 0.004 & 0.236          & 0.009 & 0.000          & 0.000 & 0.000 & 0.000 & 0.236          & 0.009 \\
200\_20\_97\_9      & 0.288          & 0.012 & 0.031 & 0.003 & 0.060 & 0.003 & 0.078 & 0.006 & \textbf{0.370} & 0.010 & 0.003          & 0.001 & 0.000 & 0.000 & \textbf{0.370} & 0.010 \\
200\_40\_130\_9\_D4 & \textbf{0.408} & 0.026 & 0.000 & 0.000 & 0.223 & 0.023 & 0.000 & 0.000 & 0.200          & 0.017 & 0.000          & 0.000 & 0.000 & 0.000 & 0.223          & 0.023 \\
200\_40\_133\_15    & \textbf{0.389} & 0.017 & 0.013 & 0.002 & 0.029 & 0.003 & 0.054 & 0.006 & 0.326          & 0.013 & 0.004          & 0.001 & 0.000 & 0.000 & 0.326          & 0.013 \\
200\_40\_45\_15     & 0.373          & 0.023 & 0.003 & 0.001 & 0.047 & 0.004 & 0.007 & 0.001 & \textbf{0.530} & 0.013 & 0.007          & 0.001 & 0.000 & 0.000 & \textbf{0.530} & 0.013 \\
200\_40\_45\_9      & 0.347          & 0.023 & 0.006 & 0.001 & 0.059 & 0.006 & 0.025 & 0.003 & \textbf{0.541} & 0.015 & 0.000          & 0.000 & 0.000 & 0.000 & \textbf{0.541} & 0.015 \\
200\_40\_90\_9      & 0.353          & 0.020 & 0.011 & 0.001 & 0.078 & 0.008 & 0.011 & 0.002 & \textbf{0.505} & 0.014 & 0.011          & 0.001 & 0.000 & 0.000 & \textbf{0.505} & 0.014 \\
200\_40\_91\_15     & 0.292          & 0.018 & 0.017 & 0.002 & 0.073 & 0.005 & 0.026 & 0.003 & \textbf{0.451} & 0.013 & 0.009          & 0.001 & 0.000 & 0.000 & \textbf{0.451} & 0.013 \\
\hline
avg                 & \textbf{0.501} & 0.031 & 0.150 & 0.010 & 0.098 & 0.009 & 0.070 & 0.007 & 0.309          & 0.015 & 0.071          & 0.010 & 0.075 & 0.009 & 0.337          & 0.016 \\
stat                        & \textbf{}                        &                         & \multicolumn{2}{l|}{\textbf{+ (P \textless 0.001)}}         & \multicolumn{2}{l|}{\textbf{+ (P \textless 0.001)}}         & \multicolumn{2}{l|}{\textbf{+ (P \textless 0.001)}}         & \multicolumn{2}{l|}{\textbf{+ (P \textless 0.001)}}         & \multicolumn{2}{l|}{\textbf{+ (P \textless 0.001)}}         & \multicolumn{2}{l||}{\textbf{+ (P \textless 0.001)}}         & \multicolumn{2}{l}{\textbf{+ (P \textless   0.005)}}
\end{tabular}
\end{adjustbox}
\end{table}

\begin{figure}[!h]
\centering
\includegraphics[width=1.0\textwidth]{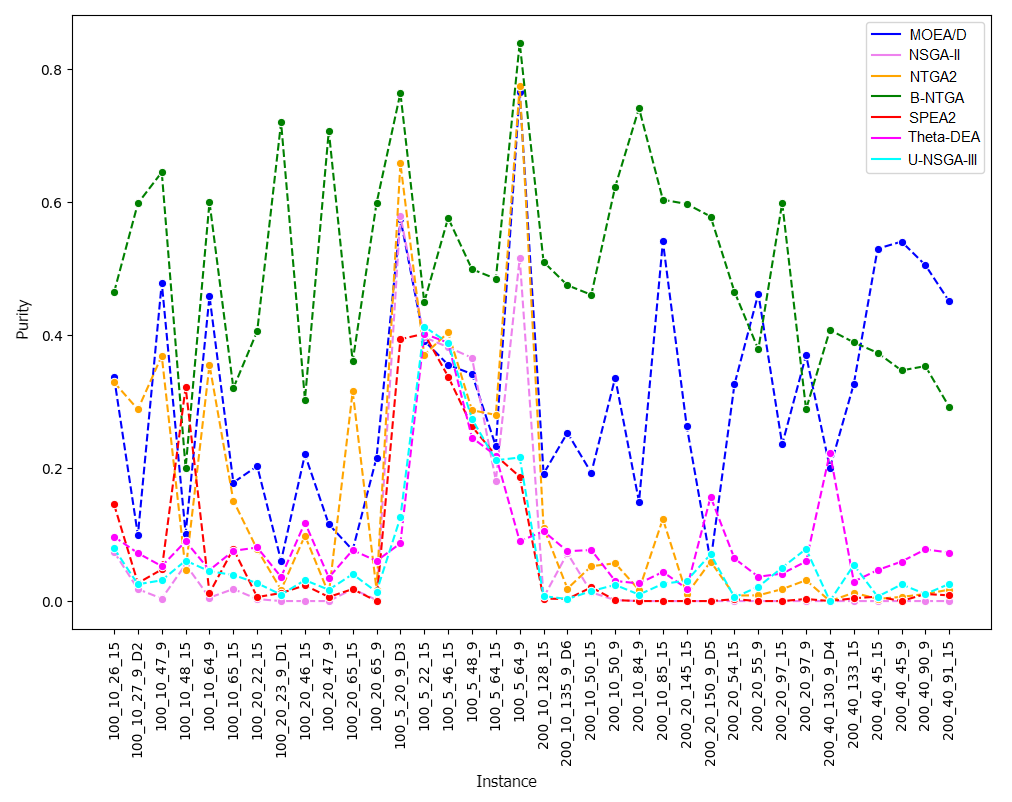}
\caption{Results for multi-objective MS-RCPSP (2obj) using \emph{Purity}}
\label{fig:msrcpsp2d-all-MND-TPFS}
\end{figure}

Fig.\ref{fig:msrcpsp2d-all-MND-TPFS} shows how examined methods cope with instances. For most instances, B-NTGA gives the best PFa (more than 50\% PFa). However, the results for four larger instances (with 200 tasks: 200\_40+) show that B-NTGA is potentially dominated by MOEA/D. However, $IGD$ (see Tab.\ref{tab:2-msrpsp-igd}) measure points out the B-NTGA. This situation is investigated in detail in many-objective MS-RCPSP experiments.    

Multi-objective MS-RCPSP that uses two objectives is a difficult-to-solve optimization problem. However, the MS-RCPSP usually is considered a many-objective problem with five objectives to examine the effectiveness of investigated methods. It makes the problem near to real-world problems but harder to solve. While each objective uses a different domain, some of the objectives conflict with each other, and the size of PFa extends tremendously.


\subsection{Results for many-objective MS-RCPSP}
\label{sub:results-5obj-MS-RCPSP}

To answer $RQ3$ question and verify how B-NTGA is effective in solving many–objective MS-RCPSP, all 5 objectives (see sec.\ref{sec:problem-deifnition}) in optimization are considered. The results of examined methods are given in Tab.\ref{tab:5-msrpsp-igd}, where IGD is measured. B-NTGA outperforms all investigated methods and gives the best PFa in each instance. 


\begin{table}[!h]
\caption{Results for many-objective MS-RCPSP (5obj) using  IGD $[10^{-3}]$
\label{tab:5-msrpsp-igd}}
\begin{adjustbox}{width=1.1\textwidth,center=\textwidth}
\begin{tabular}{l|rr|rr|rr|rr|rr|rr|rr||rr}
 & \multicolumn{2}{c|}{\textbf{B-NTGA}} & \multicolumn{2}{c|}{\textbf{NTGA2}}             & \multicolumn{2}{c|}{\textbf{$\theta$-DEA}}                 & \multicolumn{2}{c|}{\textbf{U-NSGA-III}}            & \multicolumn{2}{c|}{\textbf{MOEA/D}}                 & \multicolumn{2}{c|}{\textbf{SPEA2}}                 & \multicolumn{2}{c||}{\textbf{NSGA-II}}                & \multicolumn{2}{c}{{[}$non$ B-NTGA{]}}            \\
instance                                                          & \multicolumn{1}{c}{\textbf{avg}} & \multicolumn{1}{c|}{std} & \multicolumn{1}{c}{\textbf{avg}} & \multicolumn{1}{c|}{std} & \multicolumn{1}{c}{\textbf{avg}} & \multicolumn{1}{c|}{std} & \multicolumn{1}{c}{\textbf{avg}} & \multicolumn{1}{c|}{std} & \multicolumn{1}{c}{\textbf{avg}} & \multicolumn{1}{c|}{std} & \multicolumn{1}{c}{\textbf{avg}} & \multicolumn{1}{c|}{std} & \multicolumn{1}{c}{\textbf{avg}} & \multicolumn{1}{c||}{std} & \multicolumn{1}{c}{\textbf{avg}} & \multicolumn{1}{c}{std} \\
\hline
100\_10\_26\_15     & \textbf{0.12} & 0.01 & \textbf{0.12} & 0.01 & 0.43 & 0.06 & 0.54 & 0.10 & 0.20 & 0.01 & 1.64 & 0.10 & 0.26 & 0.01 & \textbf{0.12} & 0.01 \\
100\_10\_27\_9\_D2  & \textbf{0.13} & 0.01 & 0.20          & 0.03 & 0.64 & 0.10 & 0.79 & 0.09 & 0.31 & 0.03 & 1.83 & 0.09 & 0.34 & 0.04 & 0.20          & 0.03 \\
100\_10\_47\_9      & \textbf{0.15} & 0.02 & 0.18          & 0.02 & 0.44 & 0.04 & 0.60 & 0.04 & 0.24 & 0.02 & 1.54 & 0.09 & 0.28 & 0.01 & 0.18          & 0.02 \\
100\_10\_48\_15     & \textbf{0.08} & 0.01 & 0.09          & 0.01 & 0.29 & 0.03 & 0.38 & 0.07 & 0.18 & 0.02 & 1.23 & 0.10 & 0.14 & 0.01 & 0.09          & 0.01 \\
100\_10\_64\_9      & \textbf{0.20} & 0.02 & 0.25          & 0.03 & 0.61 & 0.06 & 0.79 & 0.07 & 0.39 & 0.03 & 1.71 & 0.08 & 0.28 & 0.02 & 0.25          & 0.03 \\
100\_10\_65\_15     & \textbf{0.12} & 0.01 & 0.14          & 0.01 & 0.44 & 0.04 & 0.52 & 0.06 & 0.25 & 0.02 & 1.71 & 0.11 & 0.28 & 0.01 & 0.14          & 0.01 \\
100\_20\_22\_15     & \textbf{0.24} & 0.02 & 0.35          & 0.03 & 0.64 & 0.04 & 0.74 & 0.07 & 0.46 & 0.03 & 1.85 & 0.12 & 0.38 & 0.02 & 0.35          & 0.03 \\
100\_20\_23\_9\_D1  & \textbf{0.31} & 0.04 & 0.43          & 0.05 & 0.79 & 0.05 & 0.91 & 0.10 & 0.58 & 0.04 & 2.11 & 0.08 & 0.53 & 0.03 & 0.43          & 0.05 \\
100\_20\_46\_15     & \textbf{0.22} & 0.01 & 0.28          & 0.02 & 0.48 & 0.04 & 0.67 & 0.06 & 0.34 & 0.02 & 1.65 & 0.10 & 0.38 & 0.01 & 0.28          & 0.02 \\
100\_20\_47\_9      & \textbf{0.26} & 0.02 & 0.39          & 0.05 & 0.86 & 0.08 & 0.96 & 0.08 & 0.51 & 0.04 & 1.76 & 0.07 & 0.52 & 0.04 & 0.39          & 0.05 \\
100\_20\_65\_15     & \textbf{0.28} & 0.01 & 0.35          & 0.04 & 0.73 & 0.08 & 0.86 & 0.08 & 0.47 & 0.04 & 2.01 & 0.07 & 0.47 & 0.03 & 0.35          & 0.04 \\
100\_20\_65\_9      & \textbf{0.27} & 0.03 & 0.34          & 0.05 & 0.70 & 0.06 & 0.90 & 0.09 & 0.58 & 0.06 & 1.84 & 0.06 & 0.37 & 0.02 & 0.34          & 0.05 \\
100\_5\_20\_9\_D3   & \textbf{0.05} & 0.01 & 0.09          & 0.03 & 0.35 & 0.02 & 0.41 & 0.05 & 0.18 & 0.02 & 0.95 & 0.09 & 0.08 & 0.00 & 0.08          & 0.00 \\
100\_5\_22\_15      & \textbf{0.08} & 0.10 & 0.20          & 0.11 & 0.11 & 0.06 & 0.33 & 0.10 & 0.10 & 0.01 & 1.00 & 0.10 & 0.45 & 0.05 & 0.10          & 0.10 \\
100\_5\_46\_15      & \textbf{0.03} & 0.00 & 0.05          & 0.00 & 0.05 & 0.00 & 0.07 & 0.01 & 0.11 & 0.01 & 0.73 & 0.11 & 0.08 & 0.00 & 0.05          & 0.00 \\
100\_5\_48\_9       & \textbf{0.05} & 0.00 & 0.09          & 0.01 & 0.09 & 0.01 & 0.19 & 0.04 & 0.12 & 0.01 & 1.06 & 0.16 & 0.10 & 0.00 & 0.09          & 0.01 \\
100\_5\_64\_15      & \textbf{0.08} & 0.01 & 0.12          & 0.01 & 0.11 & 0.02 & 0.18 & 0.02 & 0.13 & 0.01 & 1.18 & 0.10 & 0.24 & 0.01 & 0.11          & 0.02 \\
100\_5\_64\_9       & \textbf{0.11} & 0.01 & 0.14          & 0.01 & 0.30 & 0.03 & 0.40 & 0.04 & 0.20 & 0.02 & 1.15 & 0.05 & 0.35 & 0.01 & 0.14          & 0.01 \\
200\_10\_128\_15    & \textbf{0.14} & 0.03 & 0.17          & 0.02 & 0.62 & 0.08 & 0.85 & 0.08 & 0.59 & 0.05 & 1.76 & 0.09 & 0.28 & 0.01 & 0.17          & 0.02 \\
200\_10\_135\_9\_D6 & \textbf{0.24} & 0.02 & 0.32          & 0.02 & 0.70 & 0.04 & 0.76 & 0.05 & 0.35 & 0.02 & 1.51 & 0.05 & 0.42 & 0.01 & 0.32          & 0.02 \\
200\_10\_50\_15     & \textbf{0.17} & 0.02 & 0.22          & 0.02 & 0.60 & 0.05 & 0.68 & 0.08 & 0.33 & 0.02 & 1.41 & 0.10 & 0.30 & 0.01 & 0.22          & 0.02 \\
200\_10\_50\_9      & \textbf{0.16} & 0.01 & 0.24          & 0.02 & 0.56 & 0.02 & 0.65 & 0.06 & 0.27 & 0.02 & 1.18 & 0.06 & 0.30 & 0.01 & 0.24          & 0.02 \\
200\_10\_84\_9      & \textbf{0.23} & 0.03 & 0.35          & 0.03 & 0.83 & 0.03 & 0.99 & 0.07 & 0.63 & 0.04 & 1.84 & 0.10 & 0.37 & 0.01 & 0.35          & 0.03 \\
200\_10\_85\_15     & \textbf{0.08} & 0.01 & 0.13          & 0.01 & 0.69 & 0.05 & 0.77 & 0.05 & 0.32 & 0.03 & 1.33 & 0.07 & 0.14 & 0.01 & 0.13          & 0.01 \\
200\_20\_145\_15    & \textbf{0.27} & 0.03 & 0.35          & 0.03 & 0.71 & 0.06 & 0.88 & 0.05 & 0.46 & 0.03 & 1.57 & 0.06 & 0.37 & 0.01 & 0.35          & 0.03 \\
200\_20\_150\_9\_D5 & \textbf{0.43} & 0.05 & 0.58          & 0.02 & 0.66 & 0.02 & 0.82 & 0.04 & 0.44 & 0.03 & 1.50 & 0.04 & 0.46 & 0.01 & 0.44          & 0.03 \\
200\_20\_54\_15     & \textbf{0.33} & 0.04 & 0.43          & 0.04 & 0.76 & 0.04 & 0.92 & 0.03 & 0.61 & 0.05 & 1.61 & 0.08 & 0.35 & 0.01 & 0.35          & 0.01 \\
200\_20\_55\_9      & \textbf{0.35} & 0.03 & 0.53          & 0.04 & 0.93 & 0.05 & 1.10 & 0.04 & 0.70 & 0.05 & 1.79 & 0.06 & 0.44 & 0.01 & 0.44          & 0.01 \\
200\_20\_97\_15     & \textbf{0.27} & 0.02 & 0.39          & 0.04 & 0.70 & 0.05 & 0.88 & 0.08 & 0.49 & 0.04 & 1.62 & 0.06 & 0.40 & 0.01 & 0.39          & 0.04 \\
200\_20\_97\_9      & \textbf{0.29} & 0.02 & 0.46          & 0.04 & 0.83 & 0.02 & 0.98 & 0.06 & 0.52 & 0.04 & 1.73 & 0.05 & 0.41 & 0.01 & 0.41          & 0.01 \\
200\_40\_130\_9\_D4 & \textbf{0.65} & 0.07 & 1.03          & 0.06 & 1.15 & 0.06 & 1.44 & 0.08 & 1.00 & 0.05 & 2.21 & 0.04 & 0.65 & 0.02 & 0.65          & 0.02 \\
200\_40\_133\_15    & \textbf{0.39} & 0.04 & 0.57          & 0.04 & 0.79 & 0.03 & 1.02 & 0.05 & 0.55 & 0.03 & 1.66 & 0.04 & 0.56 & 0.01 & 0.55          & 0.03 \\
200\_40\_45\_15     & \textbf{0.41} & 0.03 & 0.53          & 0.04 & 0.84 & 0.06 & 1.00 & 0.05 & 0.64 & 0.03 & 1.56 & 0.03 & 0.61 & 0.01 & 0.53          & 0.04 \\
200\_40\_45\_9      & \textbf{0.43} & 0.05 & 0.52          & 0.03 & 0.81 & 0.04 & 0.97 & 0.04 & 0.63 & 0.03 & 1.62 & 0.04 & 0.70 & 0.02 & 0.52          & 0.03 \\
200\_40\_90\_9      & \textbf{0.41} & 0.02 & 0.58          & 0.03 & 0.79 & 0.06 & 0.96 & 0.05 & 0.58 & 0.02 & 1.60 & 0.05 & 0.63 & 0.02 & 0.58          & 0.03 \\
200\_40\_91\_15     & \textbf{0.38} & 0.03 & 0.55          & 0.04 & 0.83 & 0.03 & 0.99 & 0.04 & 0.64 & 0.05 & 1.60 & 0.05 & 0.52 & 0.02 & 0.52          & 0.02 \\
\hline
avg                 & \textbf{0.23} & 0.02 & 0.33          & 0.03 & 0.61 & 0.04 & 0.75 & 0.06 & 0.42 & 0.03 & 1.56 & 0.08 & 0.37 & 0.02 & 0.30          & 0.03 \\
stat                                                              & \textbf{}             &           & \multicolumn{2}{l|}{\textbf{+ (P \textless 0.001)}} & \multicolumn{2}{l|}{\textbf{+ (P \textless 0.001)}} & \multicolumn{2}{l|}{\textbf{+ (P \textless 0.001)}} & \multicolumn{2}{l|}{\textbf{+ (P \textless 0.001)}} & \multicolumn{2}{l|}{\textbf{+ (P \textless 0.001)}} & \multicolumn{2}{l||}{\textbf{+ (P \textless 0.001)}} & \multicolumn{2}{l}{\textbf{+ (P \textless 0.001)}}
\end{tabular}
\end{adjustbox}
\end{table}

The results presented in Tab.\ref{tab:5-msrpsp-mnd} show how all examined methods generate PFa. For 30/36 MS-RCPSP instances, the proposed B-NTGA gains the best average number of non-dominated solutions. The results presented in Tab.\ref{tab:5-msrpsp-igd} and Tab.\ref{tab:5-msrpsp-mnd} are statistically positively verified.

\begin{table}[!h]
\caption{Results for many-objective MS-RCPSP (5obj) using  \emph{Purity}
\label{tab:5-msrpsp-mnd}}
\begin{adjustbox}{width=1.1\textwidth,center=\textwidth}
\begin{tabular}{l|rr|rr|rr|rr|rr|rr|rr||rr}
 & \multicolumn{2}{c|}{\textbf{B-NTGA}}                          & \multicolumn{2}{c|}{\textbf{NTGA2}}                         & \multicolumn{2}{c|}{\textbf{$\theta$-DEA}}                         & \multicolumn{2}{c|}{\textbf{U-NSGA-III}}                    & \multicolumn{2}{c|}{\textbf{MOEA/D}}                                  & \multicolumn{2}{c|}{\textbf{SPEA2}}                         & \multicolumn{2}{c||}{\textbf{NSGA-II}}                              & \multicolumn{2}{c}{{[}$non$ B-NTGA{]}}                    \\
instance                    & \multicolumn{1}{c}{\textbf{avg}} & \multicolumn{1}{c|}{std} & \multicolumn{1}{c}{\textbf{avg}} & \multicolumn{1}{c|}{std} & \multicolumn{1}{c}{\textbf{avg}} & \multicolumn{1}{c|}{std} & \multicolumn{1}{c}{\textbf{avg}} & \multicolumn{1}{c|}{std} & \multicolumn{1}{c}{\textbf{avg}}      & \multicolumn{1}{c|}{std}     & \multicolumn{1}{c}{\textbf{avg}} & \multicolumn{1}{c|}{std} & \multicolumn{1}{c}{\textbf{avg}}    & \multicolumn{1}{c||}{std}    & \multicolumn{1}{c}{\textbf{avg}} & \multicolumn{1}{c}{std} \\
\hline
100\_10\_26\_15     & \textbf{0.484} & 0.008 & 0.041 & 0.002 & 0.067 & 0.002 & 0.008 & 0.001 & 0.121          & 0.001 & 0.049 & 0.000 & 0.164          & 0.002 & 0.164          & 0.002 \\
100\_10\_27\_9\_D2  & \textbf{0.472} & 0.009 & 0.034 & 0.002 & 0.036 & 0.001 & 0.003 & 0.000 & 0.148          & 0.001 & 0.038 & 0.001 & 0.233          & 0.004 & 0.233          & 0.004 \\
100\_10\_47\_9      & \textbf{0.340} & 0.006 & 0.034 & 0.002 & 0.059 & 0.003 & 0.008 & 0.000 & 0.195          & 0.001 & 0.039 & 0.000 & 0.239          & 0.004 & 0.239          & 0.004 \\
100\_10\_48\_15     & \textbf{0.376} & 0.007 & 0.037 & 0.001 & 0.035 & 0.000 & 0.009 & 0.000 & 0.099          & 0.001 & 0.024 & 0.000 & 0.215          & 0.003 & 0.215          & 0.003 \\
100\_10\_64\_9      & \textbf{0.340} & 0.007 & 0.068 & 0.004 & 0.060 & 0.002 & 0.005 & 0.001 & 0.098          & 0.001 & 0.035 & 0.001 & 0.316          & 0.004 & 0.316          & 0.004 \\
100\_10\_65\_15     & \textbf{0.461} & 0.005 & 0.040 & 0.002 & 0.075 & 0.002 & 0.012 & 0.001 & 0.158          & 0.001 & 0.049 & 0.001 & 0.148          & 0.002 & 0.158          & 0.001 \\
100\_20\_22\_15     & \textbf{0.308} & 0.005 & 0.039 & 0.002 & 0.043 & 0.003 & 0.001 & 0.000 & 0.214          & 0.002 & 0.070 & 0.001 & 0.300          & 0.006 & 0.300          & 0.006 \\
100\_20\_23\_9\_D1  & \textbf{0.390} & 0.009 & 0.056 & 0.005 & 0.038 & 0.002 & 0.001 & 0.000 & 0.173          & 0.002 & 0.063 & 0.002 & 0.261          & 0.007 & 0.261          & 0.007 \\
100\_20\_46\_15     & \textbf{0.326} & 0.006 & 0.066 & 0.004 & 0.076 & 0.005 & 0.003 & 0.000 & 0.204          & 0.002 & 0.080 & 0.001 & 0.215          & 0.005 & 0.215          & 0.005 \\
100\_20\_47\_9      & \textbf{0.386} & 0.008 & 0.046 & 0.003 & 0.035 & 0.002 & 0.001 & 0.000 & 0.190          & 0.002 & 0.053 & 0.001 & 0.269          & 0.005 & 0.269          & 0.005 \\
100\_20\_65\_15     & \textbf{0.380} & 0.009 & 0.022 & 0.003 & 0.087 & 0.012 & 0.000 & 0.000 & 0.181          & 0.002 & 0.046 & 0.002 & 0.274          & 0.007 & 0.274          & 0.007 \\
100\_20\_65\_9      & 0.268          & 0.004 & 0.018 & 0.001 & 0.098 & 0.008 & 0.003 & 0.000 & 0.164          & 0.001 & 0.065 & 0.001 & \textbf{0.357} & 0.009 & \textbf{0.357} & 0.009 \\
100\_5\_20\_9\_D3   & \textbf{0.493} & 0.009 & 0.039 & 0.002 & 0.022 & 0.001 & 0.010 & 0.000 & 0.071          & 0.000 & 0.013 & 0.000 & 0.182          & 0.002 & 0.182          & 0.002 \\
100\_5\_22\_15      & \textbf{0.574} & 0.009 & 0.080 & 0.004 & 0.063 & 0.001 & 0.048 & 0.001 & 0.045          & 0.001 & 0.027 & 0.000 & 0.017          & 0.000 & 0.080          & 0.004 \\
100\_5\_46\_15      & \textbf{0.538} & 0.006 & 0.107 & 0.001 & 0.042 & 0.001 & 0.028 & 0.001 & 0.024          & 0.000 & 0.010 & 0.000 & 0.016          & 0.000 & 0.107          & 0.001 \\
100\_5\_48\_9       & \textbf{0.543} & 0.008 & 0.041 & 0.001 & 0.046 & 0.001 & 0.013 & 0.000 & 0.051          & 0.000 & 0.010 & 0.000 & 0.056          & 0.001 & 0.056          & 0.001 \\
100\_5\_64\_15      & \textbf{0.535} & 0.007 & 0.098 & 0.005 & 0.061 & 0.001 & 0.018 & 0.001 & 0.119          & 0.001 & 0.029 & 0.001 & 0.037          & 0.001 & 0.119          & 0.001 \\
100\_5\_64\_9       & \textbf{0.408} & 0.005 & 0.026 & 0.002 & 0.106 & 0.003 & 0.028 & 0.002 & 0.103          & 0.001 & 0.050 & 0.001 & 0.097          & 0.001 & 0.106          & 0.003 \\
200\_10\_128\_15    & 0.331          & 0.007 & 0.033 & 0.001 & 0.093 & 0.004 & 0.004 & 0.000 & 0.034          & 0.000 & 0.060 & 0.001 & \textbf{0.355} & 0.004 & \textbf{0.355} & 0.004 \\
200\_10\_135\_9\_D6 & 0.125          & 0.004 & 0.044 & 0.002 & 0.091 & 0.006 & 0.000 & 0.000 & \textbf{0.354} & 0.003 & 0.025 & 0.001 & \textbf{0.355} & 0.009 & \textbf{0.355} & 0.009 \\
200\_10\_50\_15     & \textbf{0.336} & 0.008 & 0.038 & 0.003 & 0.048 & 0.002 & 0.009 & 0.001 & 0.141          & 0.001 & 0.047 & 0.000 & 0.301          & 0.005 & 0.301          & 0.005 \\
200\_10\_50\_9      & \textbf{0.282} & 0.006 & 0.031 & 0.002 & 0.093 & 0.003 & 0.006 & 0.000 & 0.212          & 0.002 & 0.034 & 0.000 & 0.237          & 0.003 & 0.237          & 0.003 \\
200\_10\_84\_9      & \textbf{0.322} & 0.006 & 0.039 & 0.002 & 0.072 & 0.003 & 0.004 & 0.000 & 0.113          & 0.001 & 0.043 & 0.000 & 0.292          & 0.004 & 0.292          & 0.004 \\
200\_10\_85\_15     & 0.264          & 0.005 & 0.021 & 0.001 & 0.043 & 0.002 & 0.010 & 0.001 & 0.076          & 0.001 & 0.025 & 0.000 & \textbf{0.466} & 0.007 & \textbf{0.466} & 0.007 \\
200\_20\_145\_15    & 0.157          & 0.004 & 0.017 & 0.001 & 0.088 & 0.006 & 0.001 & 0.000 & \textbf{0.325} & 0.002 & 0.055 & 0.001 & \textbf{0.326} & 0.008 & \textbf{0.326} & 0.008 \\
200\_20\_150\_9\_D5 & 0.112          & 0.003 & 0.025 & 0.002 & 0.070 & 0.007 & 0.000 & 0.000 & \textbf{0.479} & 0.003 & 0.040 & 0.001 & 0.267          & 0.010 & \textbf{0.479} & 0.003 \\
200\_20\_54\_15     & 0.178          & 0.005 & 0.018 & 0.001 & 0.091 & 0.005 & 0.002 & 0.000 & 0.211          & 0.001 & 0.065 & 0.001 & \textbf{0.387} & 0.007 & \textbf{0.387} & 0.007 \\
200\_20\_55\_9      & 0.235          & 0.005 & 0.033 & 0.002 & 0.095 & 0.004 & 0.002 & 0.000 & 0.208          & 0.002 & 0.051 & 0.001 & \textbf{0.308} & 0.010 & \textbf{0.308} & 0.010 \\
200\_20\_97\_15     & 0.251          & 0.006 & 0.034 & 0.001 & 0.097 & 0.007 & 0.001 & 0.000 & 0.197          & 0.002 & 0.063 & 0.001 & \textbf{0.329} & 0.010 & \textbf{0.329} & 0.010 \\
200\_20\_97\_9      & 0.151          & 0.003 & 0.038 & 0.002 & 0.090 & 0.005 & 0.002 & 0.000 & \textbf{0.313} & 0.002 & 0.051 & 0.001 & 0.290          & 0.006 & \textbf{0.313} & 0.002 \\
200\_40\_130\_9\_D4 & 0.086          & 0.003 & 0.002 & 0.000 & 0.122 & 0.014 & 0.000 & 0.000 & 0.305          & 0.003 & 0.051 & 0.002 & \textbf{0.434} & 0.018 & \textbf{0.434} & 0.018 \\
200\_40\_133\_15    & 0.147          & 0.004 & 0.031 & 0.003 & 0.098 & 0.007 & 0.000 & 0.000 & \textbf{0.397} & 0.003 & 0.039 & 0.001 & 0.259          & 0.008 & \textbf{0.397} & 0.003 \\
200\_40\_45\_15     & 0.209          & 0.005 & 0.031 & 0.002 & 0.079 & 0.005 & 0.001 & 0.000 & \textbf{0.323} & 0.003 & 0.047 & 0.001 & 0.270          & 0.007 & \textbf{0.323} & 0.003 \\
200\_40\_45\_9      & 0.253          & 0.006 & 0.039 & 0.002 & 0.080 & 0.004 & 0.002 & 0.000 & \textbf{0.311} & 0.002 & 0.049 & 0.001 & 0.210          & 0.006 & \textbf{0.311} & 0.002 \\
200\_40\_90\_9      & 0.205          & 0.005 & 0.043 & 0.003 & 0.079 & 0.006 & 0.001 & 0.000 & \textbf{0.394} & 0.002 & 0.045 & 0.001 & 0.196          & 0.008 & \textbf{0.394} & 0.002 \\
200\_40\_91\_15     & 0.202          & 0.006 & 0.021 & 0.003 & 0.100 & 0.006 & 0.001 & 0.000 & \textbf{0.342} & 0.003 & 0.038 & 0.002 & 0.256          & 0.007 & \textbf{0.342} & 0.003 \\
\hline
avg                 & \textbf{0.319} & 0.006 & 0.040 & 0.002 & 0.072 & 0.004 & 0.007 & 0.000 & 0.197          & 0.002 & 0.044 & 0.001 & 0.248          & 0.006 & 0.278          & 0.005 \\
stat                        & \textbf{}                        &                         & \multicolumn{2}{l|}{\textbf{+ (P \textless 0.001)}}         & \multicolumn{2}{l|}{\textbf{+ (P \textless 0.001)}}         & \multicolumn{2}{l|}{\textbf{+ (P \textless 0.001)}}         & \multicolumn{2}{l|}{\textbf{$\approx$ (P \textless   0.01)}} & \multicolumn{2}{l|}{\textbf{+ (P \textless 0.001)}}         & \multicolumn{2}{l||}{\textbf{$\approx$ (P \textless 0.20)}} & \multicolumn{2}{l}{\textbf{$\approx$ (P \textgreater 0.2)}}
\end{tabular}
\end{adjustbox}
\end{table}

\begin{figure}[!ht]
\centering
\includegraphics[width=1.0\textwidth]{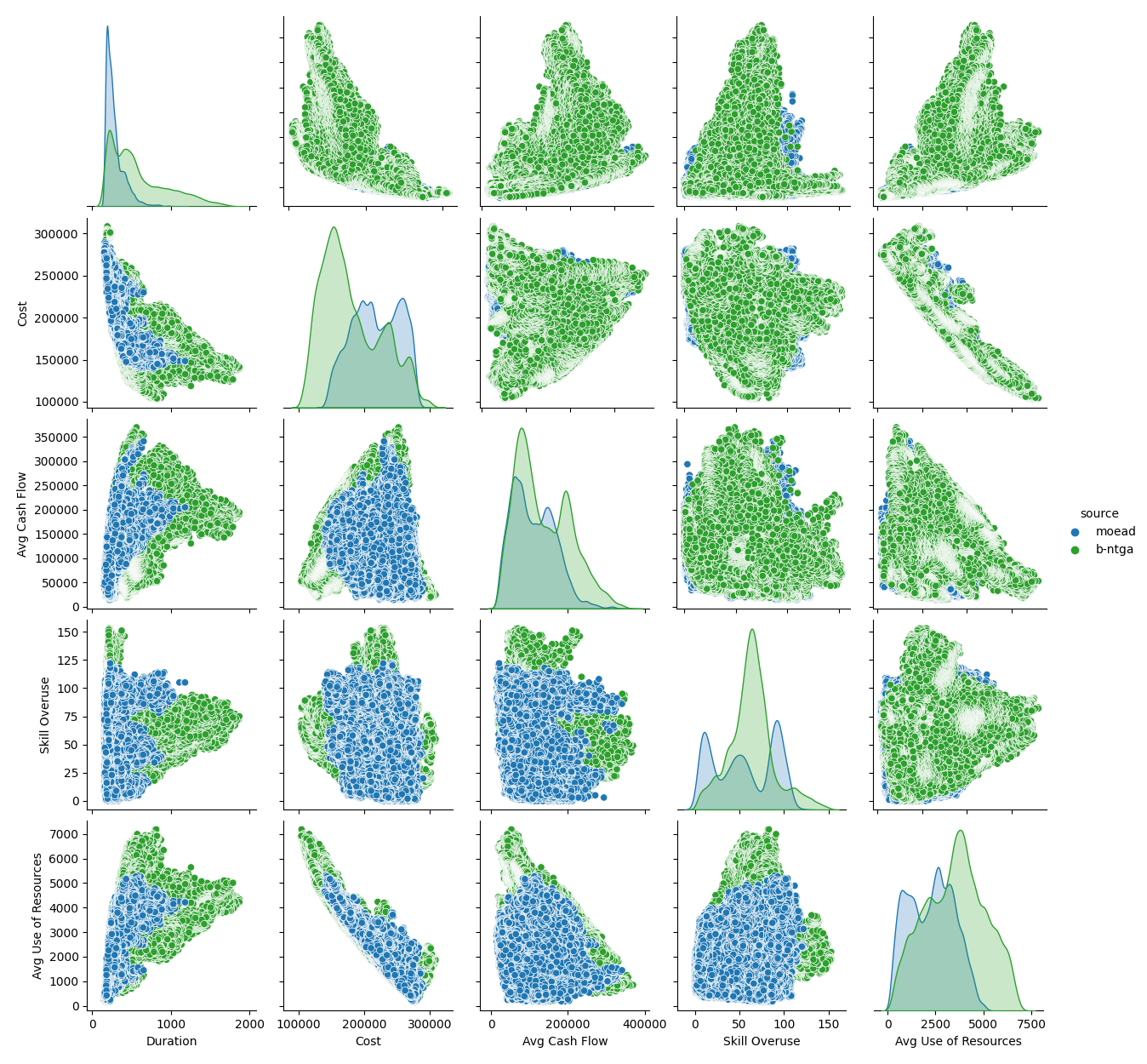}
\caption{Comparison of B-NTGA and MOEA/D Pareto Fronts Approx. for MS-RCPSP (5obj) projected onto objective pairs [instance 200\_40\_45\_9]}
\label{fig:msrcpsp5d-pf}
\end{figure}

The results in Tab.\ref{tab:5-msrpsp-mnd} show that in four bigger instances of MS-RCPSP the MOEA/D potentially outperforms other methods in terms of non-dominated solutions ($Purity$). Such a situation is in 200 task instances, almost double the results of B-NTGA. On the other hand, this trend is not observed considering $IGD$ results (see Tab.\ref{tab:5-msrpsp-igd}), suggesting those additional points are highly clustered and don't benefit the overall solution's coverage. An example is presented in Fig. \ref{fig:msrcpsp5d-pf}, where it shows B-NTGA domination as it takes a larger region of PFa.

\subsection{Summary and results discussion}
\label{sub:discussion}

Results presented in previous sections (see sec.\ref{sub:results-2obj-TTP}, \ref{sub:results-2obj-MS-RCPSP} and \ref{sub:results-5obj-MS-RCPSP}) present details of developed experiments. The additional analysis has been done to answer the last research question $RQ4$, and verify the effectiveness of B-NTGA. Results are presented in Tab.\ref{tab:summary}, where the improvement is measured by the number of non-dominated solutions found -- measured by $Purity$ and $IGD$ by each method and tested problem. B-NTGA achieved significantly better results in both multi- and many-objective of MS-RCPSP. The B-NTGA gets an average $IGD$ for the multi-objective MS-RCPSP by 52\% lower compared to all other methods combined, while for the many-objective variant, it achieved 23\% lower $IGD$ on average. The improvement is less significant for the TTP. B-NTGA outperformed $\theta$-DEA, MOEA/D, SPEA2, and NSGA-II in every instance. NTGA2 remained competitive, especially in the three smallest instances (50 items). There were other B-NTGA configurations tested during the experiments (e.g. not using \emph{infinity} value for the ‘‘edge" solutions), where performance improved on those instances. Unfortunately, the overall performance remained on a similar or lower level.

\begin{table}[!h]
\caption{Summary of averaged results for TTP and MS-RCPSP\label{tab:summary}}
\begin{adjustbox}{width=0.9\textwidth,center=\textwidth}
\begin{tabular}{l|rr|rr|rr|rr|rr|rr|rr||rr}
                                 & \multicolumn{2}{c|}{\textbf{B-NTGA}}                          & \multicolumn{2}{c|}{\textbf{NTGA2}}                         & \multicolumn{2}{c|}{\textbf{$\theta$-DEA}}                         & \multicolumn{2}{c|}{\textbf{U-NSGA-III}}                    & \multicolumn{2}{c|}{\textbf{MOEA/D}}                         & \multicolumn{2}{c|}{\textbf{SPEA2}}                         & \multicolumn{2}{c||}{\textbf{NSGA-II}}                        & \multicolumn{2}{c}{{[}$non$ B-NTGA{]}}                    \\
IGD $[10^{-3}]$ & \multicolumn{1}{c}{\textbf{avg}} & \multicolumn{1}{c|}{std} & \multicolumn{1}{c}{\textbf{avg}} & \multicolumn{1}{c|}{std} & \multicolumn{1}{c}{\textbf{avg}} & \multicolumn{1}{c|}{std} & \multicolumn{1}{c}{\textbf{avg}} & \multicolumn{1}{c|}{std} & \multicolumn{1}{c}{\textbf{avg}} & \multicolumn{1}{c|}{std} & \multicolumn{1}{c}{\textbf{avg}} & \multicolumn{1}{c|}{std} & \multicolumn{1}{c}{\textbf{avg}} & \multicolumn{1}{c||}{std} & \multicolumn{1}{c}{\textbf{avg}} & \multicolumn{1}{c}{std} \\
\hline
MS-RCPSP (2obj) & \textbf{1.54} & 0.45 & 4.07 & 0.79 & 15.23 & 1.68  & 14.15 & 1.61 & 3.77  & 0.71 & 12.59 & 1.70 & 11.74 & 1.38 & 3.22 & 1.04 \\
MS-RCPSP (5obj) & \textbf{0.23} & 0.02 & 0.33 & 0.03 & 0.61  & 0.04  & 0.75  & 0.06 & 0.42  & 0.03 & 1.56  & 0.08 & 0.37  & 0.02 & 0.30 & 0.03 \\
TTP (2obj)      & \textbf{5.52} & 2.02 & \textbf{5.63} & 1.91 & 69.96 & 11.71 & -     & -    & 11.06 & 3.82 & 17.79 & 5.75 & 14.48 & 4.07 & 5.60 & 1.90 \\
\hline
\hline
\emph{Purity}                        & \multicolumn{1}{c}{\textbf{avg}} & \multicolumn{1}{c|}{std} & \multicolumn{1}{c}{\textbf{avg}} & \multicolumn{1}{c|}{std} & \multicolumn{1}{c}{\textbf{avg}} & \multicolumn{1}{c|}{std} & \multicolumn{1}{c}{\textbf{avg}} & \multicolumn{1}{c|}{std} & \multicolumn{1}{c}{\textbf{avg}} & \multicolumn{1}{c|}{std} & \multicolumn{1}{c}{\textbf{avg}} & \multicolumn{1}{c|}{std} & \multicolumn{1}{c}{\textbf{avg}} & \multicolumn{1}{c||}{std} & \multicolumn{1}{c}{\textbf{avg}} & \multicolumn{1}{c}{std} \\
\hline
MS-RCPSP (2obj) & \textbf{0.501} & 0.031 & 0.150 & 0.010 & 0.098 & 0.009 & 0.070 & 0.007 & 0.309 & 0.015 & 0.071 & 0.010 & 0.075 & 0.009 & 0.337 & 0.016 \\
MS-RCPSP (5obj) & \textbf{0.319} & 0.006 & 0.040 & 0.002 & 0.072 & 0.004 & 0.007 & 0.000 & 0.197 & 0.002 & 0.044 & 0.001 & 0.248 & 0.006 & 0.278 & 0.005 \\
TTP (2obj)      & \textbf{0.660} & 0.084 & 0.181 & 0.025 & 0.000 & 0.000 & -     & -     & 0.098 & 0.013 & 0.016 & 0.003 & 0.012 & 0.002 & 0.238 & 0.033
\end{tabular}
\end{adjustbox}
\end{table}

The B-NTGA found (see Tab.\ref{tab:summary}) the most solutions overall, +48\% more non-dominated solutions than all other methods combined for multi-objective MS-RCPSP, +15\% for many-objective MS-RCPSP and +177\% for TTP. Despite the average results, it is worth noticing that MOEA/D especially starts to outperform other methods in terms of non-dominated solutions found in the last four (i.e. 200\_40+) instances of MS-RCPSP. However, a detailed analysis of results showed that MOEA/D focuses on some regions (i.e. clusters) of PFa that do not benefit the overall solution's coverage. 

Additional analyses have been done to verify the effectiveness of B-NTGA and compare results to reference NTGA2 method. It showed that B-NTGA compared to NTGA2, on average, gets near +10\% larger $PFa$, near 2x $Purity$ and B-NTGA uses nearly 3,5x fewer clones than NTGA2.

\begin{figure}[!h]
\centering
\includegraphics[width=1.0\textwidth]{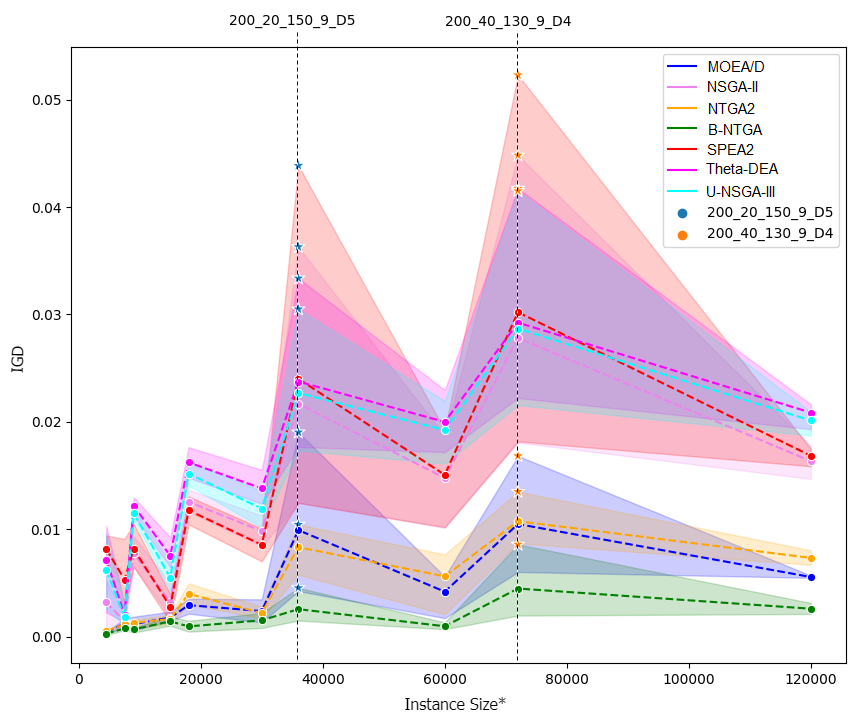}
\caption{Results summary for multi-objective MS-RCPSP (2obj) using IGD. *Instance size is approximated as [task count] * [resource count] * [skill count].}
\label{fig:msrcpsp2d-taskn-resn-skilln-IGD}
\end{figure}

\begin{figure}[!h]
\centering
\includegraphics[width=1.0\textwidth]{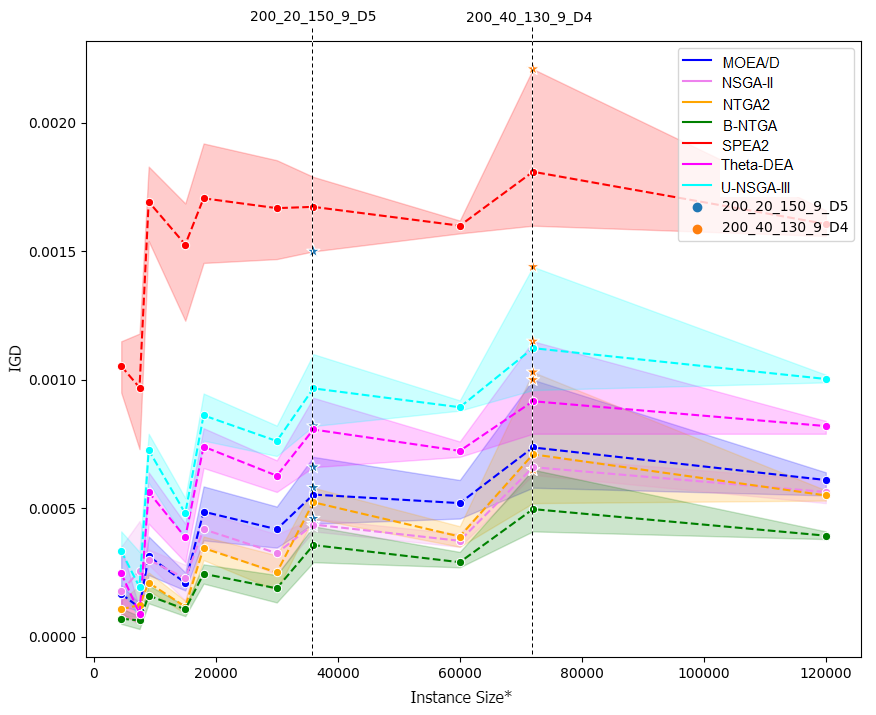}
\caption{Results summary for many-objective MS-RCPSP (5obj) using IGD. *Instance size is approximated as [task count] * [resource count] * [skill count].}
\label{fig:msrcpsp5d-taskn-resn-skilln-IGD}
\end{figure}

Moreover, to extend the $RQ4$ question we investigated a scalability factor of B-NTGA. For this experiment, we selected the MS-RCPSP problem, where there are 36 instances with various sizes and difficulties and two versions of optimization. To estimate (and measure) the difficulties of each instance, we selected a composition of the number of tasks, resources, and skills. The intuition is that the most complex instance has the larger product of the mentioned factors. Then, results of each of the examined methods (as $IGD$) have been presented -- see Fig.\ref{fig:msrcpsp2d-taskn-resn-skilln-IGD} for multi-objective MS-RCPSP and Fig.\ref{fig:msrcpsp5d-taskn-resn-skilln-IGD} for many-objective MS-RCPSP. 

A proposed instances complexity estimation (see Fig.\ref{fig:msrcpsp2d-taskn-resn-skilln-IGD} and Fig.\ref{fig:msrcpsp5d-taskn-resn-skilln-IGD}) is rather a simple one, but it is intuitive and shows a strong \emph{quasi} linear trend how methods solve instances. However, two instances break the trend -- 200\_20\_150\_9\_D5 (shortly $D5$) and 200\_40\_130\_9\_D4 (shortly $D4$), where methods get the larger standard deviation. These two instances are problematic because of many constraints, especially in many task predecessors -- in $D5$ 20.61\% tasks are on the critical path, and in $D4$ it is respectively 11.69\%. That makes instances challenging to solve. Results presented in both figures show that the proposed B-NTGA has the lowest $IGD$ values and the lowest trend for multi- and many-objective MS-RCPSP. Thus, it makes B-NTGA the most effective and scalable method too. 

Last but not least, the developed experiments showed that B-NTGA could be fragile to the instance's structure, like in MS-RCPSP $D5$ and $D4$, some 200\_40+ task instances, and for TTP instances with 50 items. In this context, despite the B-NTGA method not using any domain knowledge (i.e. specialized genetic operators), that could be an interesting future research direction.  



\section{Conclusion and Future Works}
\label{sec:conclusion}

This paper presents the B-NTGA method for multi- and many-objective optimization for two NP-hard problems (TTP and MS-RCPSP) with domain constraints. The main contribution is that the current $PFa$ (stored in \emph{archive}) in evolution's exploration/exploitation process should be guided and balanced significantly when existing constraints limit the solution landscape. Thus, in the proposed B-NTGA, the \emph{balanced} selection GAP is used. The developed research of this work is connected to five research questions to investigate B-NTGA effectiveness in the context of state-of-the-art methods and two multi- and many-objective problems. The results showed that the B-NTGA method outperforms other investigated methods. 

\todo{As proposed Pareto Front balancing mechanism of B-NTGA is generic, it can be applied to various variants of Multi-Objective Evolutionary Algorithms (MOEA), such as MOEA/D, where it can support the determination of "directions" in the decomposition vectors. It also holds substantial research potential as it allows the incorporation of modifications and enhancements developed for other domains utilizing Upper Confidence Bound (UCB), such as Monte Carlo Tree Search (MCTS). In contrast to other attempts at applying this mechanism in multi-objective optimization, the B-NTGA balancing mechanism does not necessitate the specification of a dedicated set of operators but relies solely on selection, a fundamental element of evolutionary methods. It also does not require the determination of space partitioning metrics, while still allowing their utilization if needed. 
Finally, this approach can be effectively applied to various NP-hard multi-objective combinatorial problems, especially those with many objectives. The chosen approach is also well-suited for solving problems with a multi-modal nature or deceptive characteristics} \revtwo{(see application in GaMeDE2 \cite{antkiewicz2022gamede2}}\todo{, where the optimal solution (better or different in the case of multi-modal) may lie in distinctly different spaces than the currently best solution. In particular, it could be very effective in solving optimization problems with constraints, where standard methods probe the landscape repetitively}\revtwo{, such as Capacitated Vehicle Routing Problem (CVRP) or Multi-Stage Weapon-Target Assignment Problem (MSWTA)\cite{li2015solving}.} \todo{Therefore, B-NTGA could be extended to hyper-heuristics \cite{hyper2021}.}

One interesting research direction is how to effectively manage a large/small PFa (i.e.\emph{archive}) in the B-NTGA method. If the \emph{archive} is too large, the balance procedure could get stuck in ''details''. Also, for small \emph{archive} it could work less effectively -- as near uniform distribution gives all solutions the same selection probability (see Fig.\ref{fig:ttp2d-PFa}). One potential extension of balancing GAP is \revtwo{application of clustering algorithm} to include a region counter covering solutions from the neighborhood, not only one solution.

\revtwo{One significant advantage of the B-NTGA method is its simplicity. Unfortunately, we have observed cases where excessive simplification can pose a problem. Eliminating the intermediate phase, where the selection was based on the population (as in NTGA2), generally yields better results, but theoretically, there are instances where some solutions may be unreachable. This is not possible in the examined MS-RCPSP, but it may occur in the TTP problem, where transitioning between one city permutation and another requires multiple mutation operations (reversals). A population allows for temporarily retaining dominated solutions, whereas relying solely on the archive precludes such a possibility. Determining the scenarios where a population (and its size) should be applied is a significant research direction. A potential solution could involve allowing the preservation of suboptimal solutions when the algorithm gets stuck in local optima, similar to the idea of adaptive simulated annealing.
Another instance of excessive simplification is restricting the balancing formula to the \emph{GAP} value and the number of selection attempts. While this provides a solid starting approach, it leaves much to be desired, and despite numerous attempts, expanding it has proven challenging. One tested approach was to determine the "success" of selecting a particular individual, where success meant the survival (non-domination) of the offspring or domination over existing individuals. However, such attempts did not prove effective. Additionally, favoring "edge" solutions in the Pareto front remains necessary for achieving the best results. This implies that in many cases, the same individuals are selected with a high probability of generating duplicates. Therefore, expanding on the idea of restricting the number of duplicates, a direction that should be thoroughly explored is an attempt to limit the execution of the same operations. A hypothetical solution, that will be verified in future research, is to limit the execution of the same mutation operations (at the individual or archive level — which is another aspect that needs investigation).}

Moreover, the results of experiments showed that B-NTGA effectiveness could be increased by including some problem (instance) specific knowledge. It makes B-NTGA a grey-box or glass-box optimization, but potential research directions are very promising. For example, in B-NTGA decomposition mechanisms (i.e. Dependency Structure Matrix (DSM)) usage and direct application to selection or to build new individuals (in mutation or crossover operator).
In this context, a promising future direction is linkage learning, which could join statistical and empirical linkage learning \cite{3lo} and is successfully applied to methods dedicated to solving ordering problems \cite{P4}.

The B-NTGA as the method is generic and could be applied effectively to other problems $per se$. Additionally, \emph{balanced GAP} selection could be implemented as an independent selection operator in other multi- or many-objective optimization methods. Also, the B-NTGA method could be modified by some local search algorithms, which could speed up the evolution process.

\section*{Acknowledgement}
This work was co--funded by the National Science Foundation -- 2022/45/B/ST6/04150.


\bibliographystyle{apalike}
\bibliography{article/main}

\begin{thebibliography}{}

\bibitem[Allmendinger et~al., 2022]{allmendinger2022if}
Allmendinger, R., Jaszkiewicz, A., Liefooghe, A., and Tammer, C. (2022).
\newblock What if we increase the number of objectives? theoretical and empirical implications for many-objective combinatorial optimization.
\newblock {\em Computers \& Operations Research}, 145:105857.

\bibitem[Antkiewicz et~al., 2022]{antkiewicz2022gamede2}
Antkiewicz, M., Myszkowski, P.~B., and Laszczyk, M. (2022).
\newblock Gamede2—improved gap--based memetic differential evolution applied to multimodal optimization.
\newblock In {\em 2022 17th Conference on Computer Science and Intelligence Systems (FedCSIS)}, pages 291--300. IEEE.

\bibitem[Blank et~al., 2017]{blank2017solving}
Blank, J., Deb, K., and Mostaghim, S. (2017).
\newblock Solving the bi-objective traveling thief problem with multi-objective evolutionary algorithms.
\newblock In Trautmann, H., Rudolph, G., Klamroth, K., Sch{\"u}tze, O., Wiecek, M., Jin, Y., and Grimme, C., editors, {\em Evolutionary Multi-Criterion Optimization}, pages 46--60, Cham. Springer International Publishing.

\bibitem[Bonyadi et~al., 2013]{TTP}
Bonyadi, M.~R., Michalewicz, Z., and Barone, L. (2013).
\newblock The travelling thief problem: The first step in the transition from theoretical problems to realistic problems.
\newblock In {\em 2013 IEEE Congress on Evolutionary Computation}, pages 1037--1044.

\bibitem[Browne et~al., 2012]{browne2012survey}
Browne, C.~B., Powley, E., Whitehouse, D., Lucas, S.~M., Cowling, P.~I., Rohlfshagen, P., Tavener, S., Perez, D., Samothrakis, S., and Colton, S. (2012).
\newblock A survey of monte carlo tree search methods.
\newblock {\em IEEE Transactions on Computational Intelligence and AI in games}, 4(1):1--43.

\bibitem[Das and Dennis, 1998]{ddas}
Das, I. and Dennis, J.~E. (1998).
\newblock Normal-boundary intersection: A new method for generating the pareto surface in nonlinear multicriteria optimization problems.
\newblock {\em SIAM Journal on Optimization}, 8(3):631--657.

\bibitem[Deb et~al., 2002]{nsgaii}
Deb, K., Pratap, A., Agarwal, S., and Meyarivan, T. (2002).
\newblock A fast and elitist multiobjective genetic algorithm: Nsga-ii.
\newblock {\em IEEE Transactions on Evolutionary Computation}, 6(2):182--197.

\bibitem[Dutta et~al., 2016]{ref:niching}
Dutta, S., Mallipeddi, R., and Das, K. (2016).
\newblock Seeking multiple solutions: an updated survey on niching methods and their applications.
\newblock {\em IEEE Transactions on Evolutionary Computation}, 21.6:518--538.

\bibitem[Dutta et~al., 2022]{hybrid-selection}
Dutta, S., Mallipeddi, R., and Das, K. (2022).
\newblock Hybrid selection based multi/many-objective evolutionary algorithm.
\newblock {\em Sci Rep}, 12.

\bibitem[Fortin and Marc, 2013]{ref:crowding}
Fortin, F.-A. and Marc, P. (2013).
\newblock Revisiting the nsga-ii crowding-distance computation.
\newblock {\em Proc. of the 15th annual conference on Genetic and evolutionary computation}.

\bibitem[Fu et~al., 2021]{fu2021adaptive}
Fu, G.-Z., Huang, H.-Z., Li, Y.-F., and Zhou, J. (2021).
\newblock An adaptive hybrid evolutionary algorithm and its application in aeroengine maintenance scheduling problem.
\newblock {\em Soft Computing}, 25:6527--6538.

\bibitem[Garivier and Moulines, 2011]{upper}
Garivier, A. and Moulines, E. (2011).
\newblock On upper-confidence bound policies for switching bandit problems.
\newblock In Kivinen, J., Szepesv{\'a}ri, C., Ukkonen, E., and Zeugmann, T., editors, {\em Algorithmic Learning Theory}, pages 174--188, Berlin, Heidelberg. Springer Berlin Heidelberg.

\bibitem[Hartmann and Briskorn, 2022]{HARTMANN20221}
Hartmann, S. and Briskorn, D. (2022).
\newblock An updated survey of variants and extensions of the resource-constrained project scheduling problem.
\newblock {\em European Journal of Ope. Res.}, 297(1):1--14.

\bibitem[Jian et~al., 2020]{hyper}
Jian, L., Lei, Z., and Kaizhou, G. (2020).
\newblock A genetic programming hyper-heuristic approach for the multi-skill resource constrained project scheduling problem.
\newblock {\em Expert Systems with App.}, 140:112915.

\bibitem[Laszczyk and Myszkowski, 2019]{NTGA}
Laszczyk, M. and Myszkowski, P. (2019).
\newblock Improved selection in evolutionary multi–objective optimization of multi–skill resource–constrained project scheduling problem.
\newblock {\em Information Sciences}, 481:412--431.

\bibitem[Lei et~al., 2021]{hyper2021}
Lei, Z., Jian, L., Yang-Yuan, L., and Zhou-Jing, W. (2021).
\newblock A decomposition-based multi-objective genetic programming hyper-heuristic approach for the multi-skill resource constrained project scheduling problem.
\newblock {\em Knowledge-Based Systems}, 225:107099.

\bibitem[Li et~al., 2015]{li2015solving}
Li, J., Chen, J., Xin, B., and Dou, L. (2015).
\newblock Solving multi-objective multi-stage weapon target assignment problem via adaptive nsga-ii and adaptive moea/d: A comparison study.
\newblock In {\em 2015 IEEE Congress on Evolutionary Computation (CEC)}, pages 3132--3139. IEEE.

\bibitem[Li et~al., 2013]{li2013adaptive}
Li, K., Fialho, A., Kwong, S., and Zhang, Q. (2013).
\newblock Adaptive operator selection with bandits for a multiobjective evolutionary algorithm based on decomposition.
\newblock {\em IEEE Transactions on Evolutionary Computation}, 18(1):114--130.

\bibitem[Li, 2005]{ref:speciation}
Li, X. (2005).
\newblock Efficient differential evolution using speciation for multimodal function optimization.
\newblock {\em Proc. of the 7th annual conference on Genetic and evolutionary computation}.

\bibitem[Ling and Xiao-long, 2018]{fruit-mmo}
Ling, W. and Xiao-long, Z. (2018).
\newblock A knowledge-guided multi-objective fruit fly optimization algorithm for the multi-skill resource constrained project scheduling problem.
\newblock {\em Swarm and Evolutionary Computation}, 38:54--63.

\bibitem[Luiz Gustavo~Dias et~al., 2016]{referepoints}
Luiz Gustavo~Dias, L., Tarc{\'i}sio, G.~B., Anderson, P. d.~P., Rog{\'e}rio, S.~P., and Pedro~Paulo, B. (2016).
\newblock Robust parameter optimization based on multivariate normal boundary intersection.
\newblock {\em Comput. Ind. Eng.}, 93:55--66.

\bibitem[Myszkowski and et~al., 2023]{ref:imopse}
Myszkowski, P. and et~al. (2023).
\newblock imopse project, http://imopse.ii.pwr.edu.pl.

\bibitem[Myszkowski and Laszczyk, 2019]{Survey}
Myszkowski, P. and Laszczyk, M. (2019).
\newblock Survey of quality measures for multi-objective optimization: Construction of complementary set of multi-objective quality measures.
\newblock {\em Swarm and Evolutionary Computation}, 48:109--133.

\bibitem[Myszkowski and Laszczyk, 2021]{NTGA2}
Myszkowski, P. and Laszczyk, M. (2021).
\newblock Diversity based selection for many-objective evolutionary optimisation problems with constraints.
\newblock {\em Information Sciences}, 546:665--700.

\bibitem[Myszkowski and Laszczyk, 2022]{benchmark2022}
Myszkowski, P. and Laszczyk, M. (2022).
\newblock Investigation of benchmark dataset for many-objective multi-skill resource constrained project scheduling problem.
\newblock {\em Applied Soft Computing}, 127:109253.

\bibitem[Myszkowski et~al., 2018]{degr}
Myszkowski, P., Olech, L., Laszczyk, M., and Skowronski, M. (2018).
\newblock Hybrid differential evolution and greedy algorithm (degr) for solving multi-skill resource-constrained project scheduling problem.
\newblock {\em Applied Soft Computing}, 62:1--14.

\bibitem[Myszkowski et~al., 2015a]{benchmar2015}
Myszkowski, P., Skowronski, M.~E., and Sikora, K. (2015a).
\newblock A new benchmark dataset for multi-skill resource-constrained project scheduling problem.
\newblock In {\em 2015 Federated Conf. on Computer Science and Information Systems (FedCSIS)}, pages 129--138.

\bibitem[Myszkowski et~al., 2015b]{hantco}
Myszkowski, P.~B., Skowronski, M., Olech, L., and Oslizlo, K. (2015b).
\newblock Hybrid ant colony optimization in solving multi-skill resource-constrained project scheduling problem.
\newblock {\em Soft Comput}, 19.

\bibitem[Polyakovskiy et~al., 2014]{ttp-inst}
Polyakovskiy, S., Bonyadi, M., Wagner, M., Michalewicz, Z., and Neumann, F. (2014).
\newblock A comprehensive benchmark set and heuristics for the traveling thief problem.

\bibitem[Przewozniczek and Komarnicki, 2020]{3lo}
Przewozniczek, M.~W. and Komarnicki, M.~M. (2020).
\newblock Empirical linkage learning.
\newblock {\em IEEE Transactions on Evolutionary Computation}, 24(6):1097--1111.

\bibitem[Seada and Kalyanmoy, 2015]{u-nsgaiii}
Seada, H. and Kalyanmoy, D. (2015).
\newblock U-nsga-iii: A unified evolutionary algorithm for single, multiple, and many-objective optimization.
\newblock {\em COIN report}, 2014022.

\bibitem[Sfikas et~al., 2021]{sfikas2021monte}
Sfikas, K., Liapis, A., and Yannakakis, G.~N. (2021).
\newblock Monte carlo elites: Quality-diversity selection as a multi-armed bandit problem.
\newblock In {\em Proceedings of the Genetic and Evolutionary Computation Conference}, pages 180--188.

\bibitem[Sun and Li, 2020]{sun2020adaptive}
Sun, L. and Li, K. (2020).
\newblock Adaptive operator selection based on dynamic thompson sampling for moea/d.
\newblock In {\em International Conference on Parallel Problem Solving from Nature}, pages 271--284. Springer.

\bibitem[Verma et~al., 2021]{snasel-surv}
Verma, S., Pant, M., and Snasel, V. (2021).
\newblock A comprehensive review on nsga-ii for multi-objective combinatorial optimization problems.
\newblock {\em IEEE Access}, 9:57757--57791.

\bibitem[Vijayan and et~al., 1992]{nair1992taguchi}
Vijayan, N.~N. and et~al. (1992).
\newblock Taguchi's parameter design: A panel discussion.
\newblock {\em Technometrics}, 34(2):127--161.

\bibitem[Wozniak et~al., 2020]{P4}
Wozniak, S., Przewozniczek, M.~W., and Komarnicki, M.~M. (2020).
\newblock Parameter-less population pyramid for permutation-based problems.
\newblock In {\em Parallel Problem Solving from Nature – PPSN XVI: 16th International Conference, PPSN 2020, Leiden, The Netherlands, September 5-9, 2020, Proceedings, Part I}, page 418–430, Berlin, Heidelberg. Springer-Verlag.

\bibitem[Xiaolong et~al., 2015]{fruit}
Xiaolong, Z., Ling, W., and Huanyu, Z. (2015).
\newblock A knowledge-based fruit fly optimization algorithm for multi-skill resource-constrained project scheduling problem.
\newblock In {\em 2015 34th Chinese Control Conference (CCC)}, pages 2615--2620.

\bibitem[Yuan and et~al., 2015]{thetadea}
Yuan, Y. and et~al. (2015).
\newblock A new dominance relation-based evolutionary algorithm for many-objective optimization.
\newblock {\em IEEE Transactions on Evolutionary Computation}, 20(1).

\bibitem[Zhang and Li, 2007]{moead}
Zhang, Q. and Li, H. (2007).
\newblock Moea/d: A multiobjective evolutionary algorithm based on decomposition.
\newblock {\em IEEE Transactions on Evolutionary Computation}, 11(6):712--731.

\bibitem[Zheng et~al., 2017]{tlbo}
Zheng, H., Wang, L., and Zheng, X. (2017).
\newblock Teaching–learning-based optimization algorithm for multi-skill resource constrained project scheduling problem.
\newblock {\em Soft Computing}, 21:1537–1548.

\bibitem[Zitzler et~al., 2001]{spea2}
Zitzler, E., Laumanns, M., and Thiele, L. (2001).
\newblock Spea2: Improving the strength pareto evolutionary algorithm.
\newblock {\em ETH Zurich, Computer Eng. and Networks Lab.}, 103.

\end{thebibliography}

\end{document}